\documentclass{article} % For LaTeX2e
\usepackage{iclr2025_conference,times}
\usepackage{graphicx}
\usepackage{subcaption}
\usepackage{caption}
\usepackage{booktabs}
\usepackage{wrapfig}

\usepackage{amsmath,amsfonts,bm}

\def\eqref#1{equation~\ref{#1}}
\def\1{\bm{1}}

\def\vd{{\bm{d}}}

\def\vm{{\bm{m}}}

\DeclareMathAlphabet{\mathsfit}{\encodingdefault}{\sfdefault}{m}{sl}
\SetMathAlphabet{\mathsfit}{bold}{\encodingdefault}{\sfdefault}{bx}{n}

\usepackage{hyperref}
\usepackage{url}
\definecolor{citecolor}{HTML}{0071BC}
\definecolor{linkcolor}{HTML}{1e0afc}
\hypersetup{%
  pdfborder=0 0 0,
  colorlinks=true,
  citecolor=citecolor,
}

\title{Scaling Properties of Diffusion Models \\ for Perceptual Tasks}
\author{Rahul Ravishankar*, Zeeshan Patel*,  Jathushan Rajasegaran, Jitendra Malik \\
University of California, Berkeley \\
\texttt{\{rravishankar, zeeshanp\}@berkeley.edu} \\
}

\iclrfinalcopy % Uncomment for camera-ready version, but NOT for submission.
\begin{document}

\maketitle

\def\thefootnote{*}\footnotetext{Equal Contribution}\def\thefootnote{\arabic{footnote}}

\begin{abstract}

In this paper, we argue that iterative computation with diffusion models offers a powerful paradigm for not only generation but also visual perception tasks. We unify tasks such as depth estimation, optical flow, and amodal segmentation under the framework of image-to-image translation, and show how diffusion models benefit from scaling training and test-time compute for these perceptual tasks. Through a careful analysis of these scaling properties, we formulate compute-optimal training and inference recipes to scale diffusion models for visual perception tasks. Our models achieve competitive performance to state-of-the-art methods using significantly less data and compute. We release code and models at \href{https://scaling-diffusion-perception.github.io}{scaling-diffusion-perception.github.io}.

\end{abstract}

% \vspace{0.2cm}
\section{Introduction}
% \vspace{0.2cm}

\begin{figure*}[htbp]
    \centering
    \includegraphics[width=0.9\linewidth]{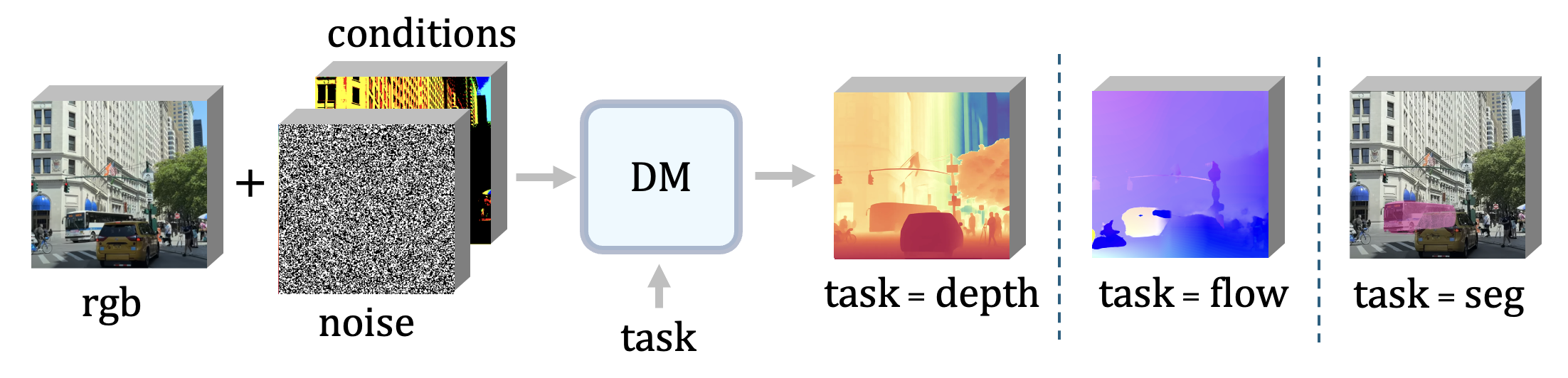}
    \small
    
    \caption{\textbf{A Unified Framework:} We fine-tune a pre-trained Diffusion Model (DM), for visual perception tasks. We take a RGB image, and a conditional image (i.e. next video frame, occlusion mask, etc.), along with the noised image of the ground truth prediction. Our model generates predictions for visual tasks such as depth estimation, optical flow prediction, and amodal segmentation, based on the conditional task embedding. We train a generalist model that can perform all three tasks with exceptional performance.}
    \label{fig:main_fig}
    \vspace{-0.4cm}
\end{figure*}

Diffusion models have emerged as powerful techniques for generating images and videos, while showing excellent scaling behaviors. In this paper, we present a unified framework to perform a variety of perceptual tasks --- depth estimation, optical flow estimation, and amodal segmentation --- with a single diffusion model, as illustrated in Figure~\ref{fig:main_fig}. 
%In this paper, we present a unified framework to train a single generalist diffusion model to solve various perceptual tasks, as illustrated in Figure~\ref{fig:main_fig}.

Previous works such as Marigold~\citep{ke2024repurposing}, FlowDiffuser~\citep{luo2024flowdiffuser}, and pix2gestalt~\citep{ozguroglu2024pix2gestalt} demonstrate the potential of repurposing image diffusion models for various inverse vision tasks individually. Building on these prior works, we perform an extensive empirical study, establishing scaling power laws for depth estimation, and display their transferability to other perceptual tasks. Using insights from these scaling laws, we formulate compute-optimal recipes for diffusion training and inference. We find that efficiently scaling compute for diffusion models leads to significant performance gains in downstream perceptual tasks.

% Previous works such as Marigold~\citep{ke2024repurposing}, FlowDiffuser~\citep{luo2024flowdiffuser}, and pix2gestalt~\citep{ozguroglu2024pix2gestalt} repurpose image diffusion models for depth estimation, amodal segmentation, and optical flow estimation, respectively. Through an extensive empirical study, we establish scaling power laws for these inverse vision problems. Our work is the first to show that efficiently scaling compute for diffusion models significantly improves downstream performance on perceptual tasks.

Recent works in other fields have also focused on scaling test-time compute to enhance the capabilities of modern LLMs, as demonstrated by OpenAI's o1 model~\citep{openai2024o1}. Noam Brown, one of the key authors, expressed it quite pithily in a Ted Talk, \emph{``It turned out that having a bot think for just 20 seconds in a hand of poker got the same boosting performance as scaling up the model by 100,000x and training it for 100,000 times longer.''} In our experiments, we observe a similar trade off between allocating more compute during training versus test-time for diffusion models with respect to downstream performance on perceptual tasks. 

We scale test-time compute by exploiting the iterative and stochastic nature of diffusion to increase the number of denoising steps. By allocating more compute to early denoising steps, and ensembling multiple denoised predictions, we consistently achieve higher accuracy on these perceptual tasks. Our results provide evidence of the benefits of scaling test-time compute for inverse vision problems under constrained compute budgets, bringing a new perspective to the conventional paradigm of training-centric scaling for generative models.

\section{Related Work}

\textbf{Generative Modeling:} Generative modeling has been studied under various methods, including VAEs~\citep{kingma2013auto}, GANs~\citep{goodfellow2014generative}, Normalizing Flows~\citep{rezende2015variational}, Autoregressive models~\citep{pmlr-v48-oord16}, and Diffusion models~\citep{sohl2015deep, ho2020denoising}. Denoising Diffusion Probabilistic Models (DDPMs)~\citep{ho2020denoising} have shown impressive scaling behaviors for many image and video generation models. Notable examples include Latent Diffusion Models~\citep{rombach2022high}, which enhanced efficiency by operating in a compressed latent space, Imagen~\citep{saharia2022photorealistic}, which generates samples in pixel space with increasing resolution, and Consistency Models~\citep{song2023consistency}, which aim to accelerate sampling while maintaining generation quality. Recent methods like Rectified Flow~\citep{liu2022flowmatching} and Flow Matching~\citep{lipman2023flowmatching} employ training objectives inspired by optimal transport to model continuous vector fields that map data to target distributions, eliminating the discrete formulation of diffusion models.
Rectified Flow mitigates numerical issues in training by applying flow regularization, and Flow Matching offers efficient sampling with fewer discretization artifacts, making them promising alternatives to diffusion for high-quality generation. 
Apart from diffusion models, Parti~\citep{yu2022scaling} and MARS~\citep{he2024marsmixtureautoregressivemodels} showcased the potential of autoregressive models for image generation, and the Muse architecture~\citep{chang2023musetexttoimagegenerationmasked} introduced a masked image generation approach using transformers.

% scaling with diffusion
\textbf{Scaling Diffusion Models:} Diffusion modeling has shown impressive scaling behaviors in terms of data, model size, and compute. Latent Diffusion Models~\citep{rombach2022high} first showed that training with large-scale web datasets can achieve high quality image generation results with a U-Net backbone. DiT~\citep{peebles2023scalable} explored scaling diffusion models with the transformer architecture, presenting desirable scaling properties for class-conditional image generation. Later, \emph{Li et al.}\citep{li2024scalability} studied alignment scaling laws of text-to-image diffusion models. Recently, \emph{Fei et al.}\citep{fei2024scaling} trained mixture-of-experts DiT models up to 16B parameters, achieving high-quality image generation results. Upcycling is another way to scale transformer models. \emph{Komatsuzaki et al.}~\citep{komatsuzaki2022sparse} used upcycling to convert a dense transformer-based language model to a mixture-of-experts model without pre-training from scratch. Similarly, EC-DiT\cite{sun2024ecditscalingdiffusiontransformers} explores how to exploit heterogeneous compute allocation in mixture-of-experts training for DiT models through expert-choice routing and learning to adaptively optimize the compute allocated to specific text-image data samples.

% for inverse problems
\textbf{Diffusion Models for Perception Tasks:} Diffusion models have also been used for various downstream visual tasks such as depth estimation~\citep{ji2023ddp, duan2023diffusiondepth, saxena2023monocular, saxena2024surprising, zhao2023unleashing}. More recently, Marigold~\citep{ke2024repurposing} and GeoWizard~\citep{fu2024geowizardunleashingdiffusionpriors} displayed impressive results by repurposing pre-trained diffusion models for monocular depth estimation. Diffusion models with few modifications are used for semantic segmentation for categorical distributions~\citep{hoogeboom2021argmax, brempong2022denoising, tan2022semantic, amit2021segdiff, baranchuk2021label, wolleb2022diffusion}, instance segmentation~\citep{gu2024diffusioninst}, and panoptic segmentation~\citep{chen2023generalist}. Diffusion models are also used for optical flow~\citep{luo2024flowdiffuser, saxena2024surprising} and 3D understanding~\citep{liu2023zero, jain2022zero, poole2022dreamfusion, wang2023score, watson2022novel}.

\section{Generative Pre-Training}

We first explore how to efficiently scale diffusion model pre-training. We pre-train diffusion models for class-conditional image generation using a diffusion transformer (DiT) backbone and follow the original model training recipe~\citep{peebles2023scalable}.

Starting with a target RGB image \(I \in \mathbb{R}^{u \times u \times 3}\), where the resolution of the image is $u \times u$, our pretrained, frozen Stable Diffusion variational autoencoder \citep{rombach2022high} compresses the target to a latent \(z_0 \in \mathbb{R}^{w \times w \times 4}\), where $w = u / 8$. Gaussian noise is added at sampled time steps to obtain a noisy target latent. Noisy samples are generated as:
\begin{equation}
z_t = \sqrt{\alpha_t} \cdot z_0 + \sqrt{1 - \alpha_t} \cdot \epsilon_t 
\end{equation}
for timestep \(t\).
The noise is distributed as \(\epsilon \sim \mathcal{N}(0, I)\), \(t \sim \text{Uniform}(T)\), with \(T = 1000\) and \(\alpha_t := \prod_{s=1}^t (1-\beta_s)\), with \(\{\beta_1, \dots, \beta_T\}\) as the variance schedule of a process.

In the denoising process, the class-conditional DiT  \(f_\theta(\cdot)\), parameterized by learned parameters \(\theta\), gradually removes noise from \(z_t\) to obtain \(z_{t-1}\). The parameters \(\theta\) are updated by noising \(z_0\) with sampled noise \(\epsilon\) at a random timestep \(t\), computing the noise estimate, and optimizing the mean squared loss between the generated noise and estimated noise in an \(n\) batch size sample. We formally represent this as the following minimization problem:
\begin{equation}
\theta^* = \arg\min_\theta \mathcal{L}_\theta(z_t, \epsilon_i) = \arg\min_\theta \frac{1}{n} \sum_{i=1}^n (\epsilon_i - \hat{\epsilon}_i)^2,
\end{equation}
where $\theta^*$ are the DiT learned parameters and $\hat{\epsilon}_i$ is the DiT noise prediction for sample $i$.

\subsection{Model Size}
\label{sec:model_size_pretrain}
\begin{wrapfigure}{r}{0.5\textwidth}
    \centering
    \vspace{-1cm}
    \includegraphics[width=0.99\linewidth]{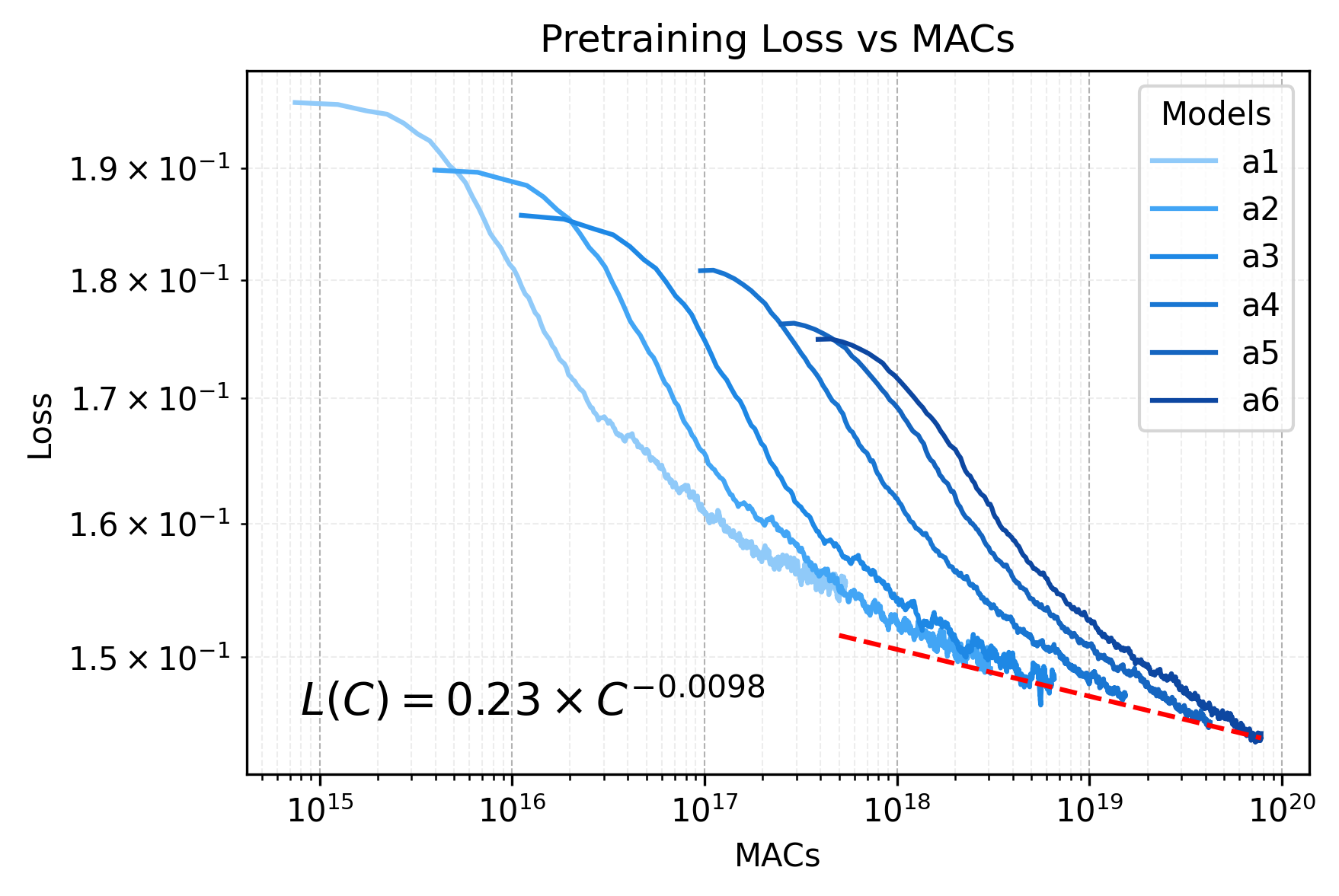}
    \caption{\textbf{Scaling at Model Size:} For generative pre-training of DiT models, we observe clear power law scaling behavior as we increase the model size.}
    \label{fig:a1_a6_loss}
    \vspace{-0.5cm}
\end{wrapfigure}

We pre-train six different dense DiT models as in Table~\ref{tab:dense_dit_model}, increasing model size by varying the number of layers and hidden dimension size. We use Imagenet-1K~\citep{russakovsky2015imagenetlargescalevisual} as our pre-training dataset and train all models for 400k iterations with a fixed learning rate of $1e$-$4$ and a batch size of 256. Fig.~\ref{fig:a1_a6_loss} shows that larger models converge to lower loss with a clear power law scaling behavior. We show the train loss as a function of compute (in MACs), and our predictions indicate a power law relationship of $L(C) = 0.23 \times C^{-0.0098}$. Our pre-training experiments display the ease of scaling DiT with a small training dataset, which translates directly to efficiently scaling downstream model performance.

\subsection{Mixture of Experts}
We also pre-train Sparse Mixture of Experts (MoE) models \citep{shazeer2017outrageouslylargeneuralnetworks}, following the S/2 and L/2 model configurations in \citep{FeiDiTMoE2024}. We use three different MoE configurations listed in Table \ref{tab:moe_dit_models}, scaling the total parameter count by increasing hidden size, number of experts, layers, and attention heads. Each MoE block activates the top-2 experts per token and has a shared expert that is used by all tokens. 
To alleviate issues with expert balance, we use the proposed expert balance loss function from \citep{FeiDiTMoE2024} which distributes the load across experts more efficiently. Sparse MoE pre-training allows for a higher parameter count while increasing throughput, making it more compute efficient than training a dense DiT model of the same size. We train our DiT-MoE models with the same training recipe as the dense DiT model using ImageNet-1K. Our approach enables training DiT-MoE models to increase model capacity without increasing compute usage by another order of magnitude, which would be required to train dense models of similar sizes.

%##################################################################################################
\begin{table}[!t]
\begin{center}
\footnotesize
\setlength{\tabcolsep}{4pt}
\begin{minipage}{0.48\textwidth}
\begin{tabular}{c c c c c} 
\toprule[0.4mm]
\textbf{Model} & \textbf{Params} & \textbf{Dimension} & \textbf{Heads} & \textbf{Layers} \\ \midrule
a1  & 14.8M & 256  & 16 & 12 \\
a2 & 77.2M & 512 & 16 & 16 \\
a3    & 215M & 768 & 16 & 20 \\
a4    & 458M & 1024 & 16 & 24 \\
a5    & 1.2B & 1536 & 16 & 28 \\
a6    & 1.9B & 1792 & 16 & 32 \\ 
\bottomrule[0.4mm]
\end{tabular}
\caption{\textbf{Dense DiT Models:} We scale dense DiT model size by increasing hidden dimension and number of layers linearly while keeping number of heads constant following \citep{yang2022tensorprogramsvtuning, touvron2023llama2openfoundation, }.}
\label{tab:dense_dit_model}
\end{minipage}
\hfill
\begin{minipage}{0.48\textwidth}
\setlength{\tabcolsep}{3pt}
\begin{tabular}{c c c c c} 
% \multicolumn{6}{c}{~}\\
\toprule[0.4mm]
\textbf{Model} & \textbf{Active / Total} & \textbf{Dim} & \textbf{Heads} & \textbf{Layers} \\ \midrule
S/2-8E2A  & 71M / 199M  & 384  & 6 & 12 \\
S/2-16E2A & 71M / 369M & 384 & 6 & 12 \\
L/2-8E2A    & 1.0B / 2.8B & 1024 & 16 & 24 \\
\bottomrule[0.4mm]
\multicolumn{5}{c}{~}\\
% \multicolumn{6}{c}{~}\\
\end{tabular}
\vspace{0.25cm}
\caption{\textbf{MoE DiT Models:} We scale the MoE DiT models by increasing dimension size, number attention heads, layers, and experts following \citep{FeiDiTMoE2024}.}
\label{tab:moe_dit_models}
\end{minipage}
\end{center}
\end{table}
%##################################################################################################

\section{Fine-Tuning for Perceptual tasks}

% unification of all tasks, need to explain more
In this section, we explore how to scale the fine-tuning of the pre-trained DiT models to maximize performance on downstream perception tasks. During fine-tuning, we utilize the image-to-image diffusion process from ~\citep{ke2024repurposing} and ~\citep{brooks2023instructpix2pixlearningfollowimage} as our training recipe. We pose all our visual tasks as conditional denoising diffusion generation. Give an RGB image \(I \in \mathbb{R}^{u \times u \times 3}\) and its pair ground truth image \(D \in \mathbb{R}^{u \times u \times 3}\), we first project them to the latent space, \(i_0 \in \mathbb{R}^{w \times w \times 4}\) and \(d_0 \in \mathbb{R}^{w \times w \times 4}\), respectively. We only add noise to the ground truth latent to obtain \(d_t\) and concatenate it with the RGB latent which results in a tensor \(z_t = \{i_0, d_t\}\). The first convolutional layer of the DiT model is modified to match the doubled number of input channels, and its values are reduced by half to make sure the predictions are the same if the inputs are just RGB images~\citep{ke2024repurposing}. Finally, we perform diffusion training by denoising the ground truth image. We ablate several fine-tuning compute scaling techniques on the monocular depth estimation task and report Absolute Relative error and Delta1 error. We transfer the best configurations from the depth estimation ablation study to fine-tune for other visual perception tasks.

\subsection{Effect of Model Size}
\label{sec:finetune}

\begin{figure}[htbp]
    \centering
    \begin{minipage}[b]{0.31\textwidth}
        \centering
        \includegraphics[width=0.99\linewidth]{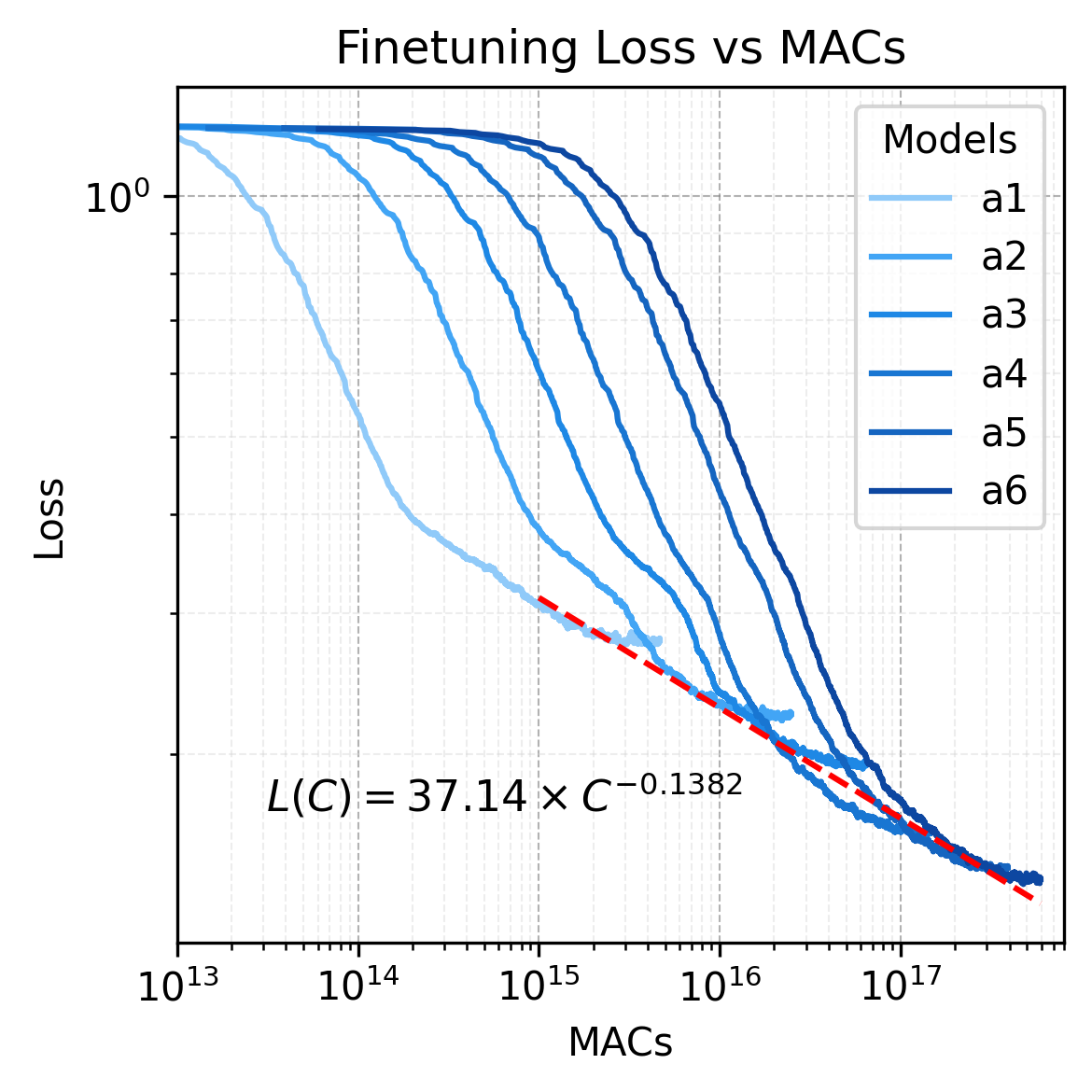}
        % \caption*{(a)}
    \end{minipage}
    \hfill
    \begin{minipage}[b]{0.31\textwidth}
        \centering
        \includegraphics[width=0.99\linewidth]{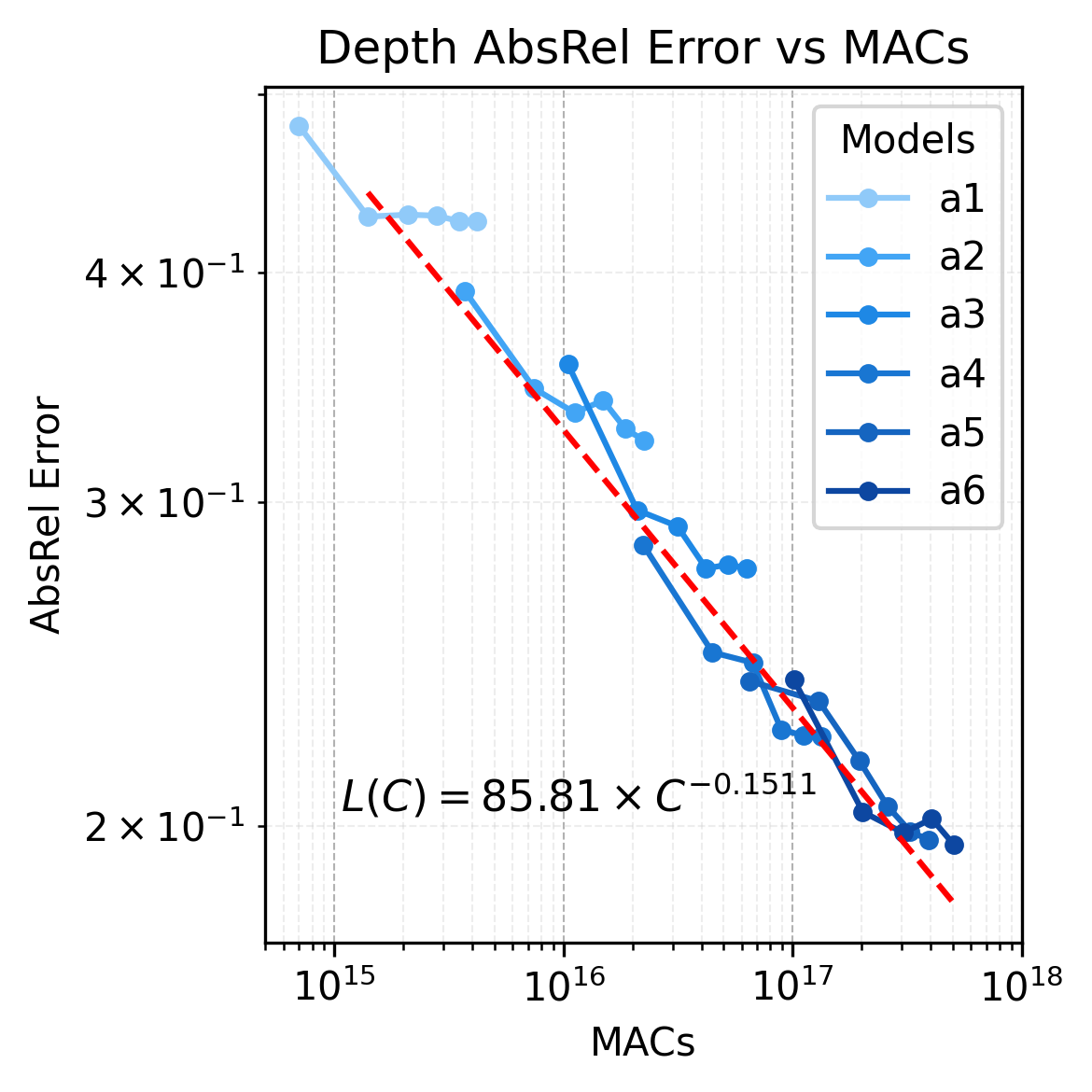}
        % \caption*{(b)}
    \end{minipage}
    \hfill
    \begin{minipage}[b]{0.31\textwidth}
        \centering
        \includegraphics[width=0.99\linewidth]{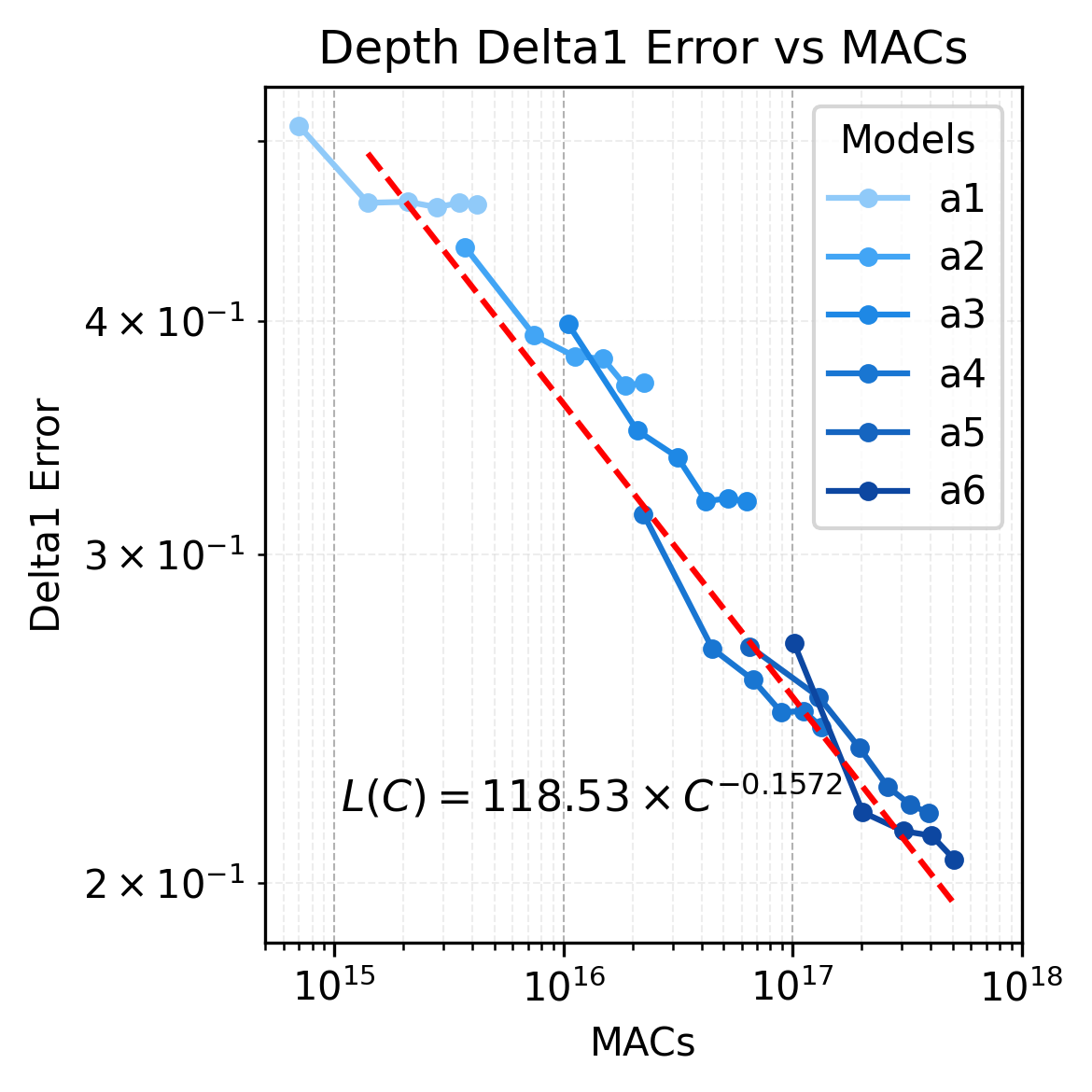}
        % \caption*{(c)}
    \end{minipage}
    \caption{\textbf{Effect of Model Size:} We fine-tune a1-a6 models on the Hypersim dataset for 30K iterations with an exponential decay learning rate schedule from $3e$-$5$ to $3e$-$7$. We observe a strong correlation between the fine-tuning loss scaling law and validation metric scaling laws.}
    \label{fig:dense_finetune_size_scaling}
\end{figure}

We fine-tune the pre-trained a1-a6 dense models on the depth estimation task to study the effect of model size. We scale model size as shown in as described in Section~\ref{sec:model_size_pretrain}. Fig.~\ref{fig:dense_finetune_size_scaling} shows that larger dense DiT models predictably converge to a lower fine-tuning loss, presenting a clear power law scaling behavior. We plot the train loss and validation metrics as a function of compute (in MACs). Our fine-tuned model predictions show a power law relationship in both depth Absolute Relative error and depth Delta1 error. These experiments provide strong signal on how model performance will scale as we increase fine-tuning compute by scaling model size.

\subsection{Effect of Pre-training Compute}
\begin{figure}[htbp]
    \centering
    \begin{minipage}[b]{0.48\textwidth}
        \centering
        \includegraphics[width=0.99\linewidth]{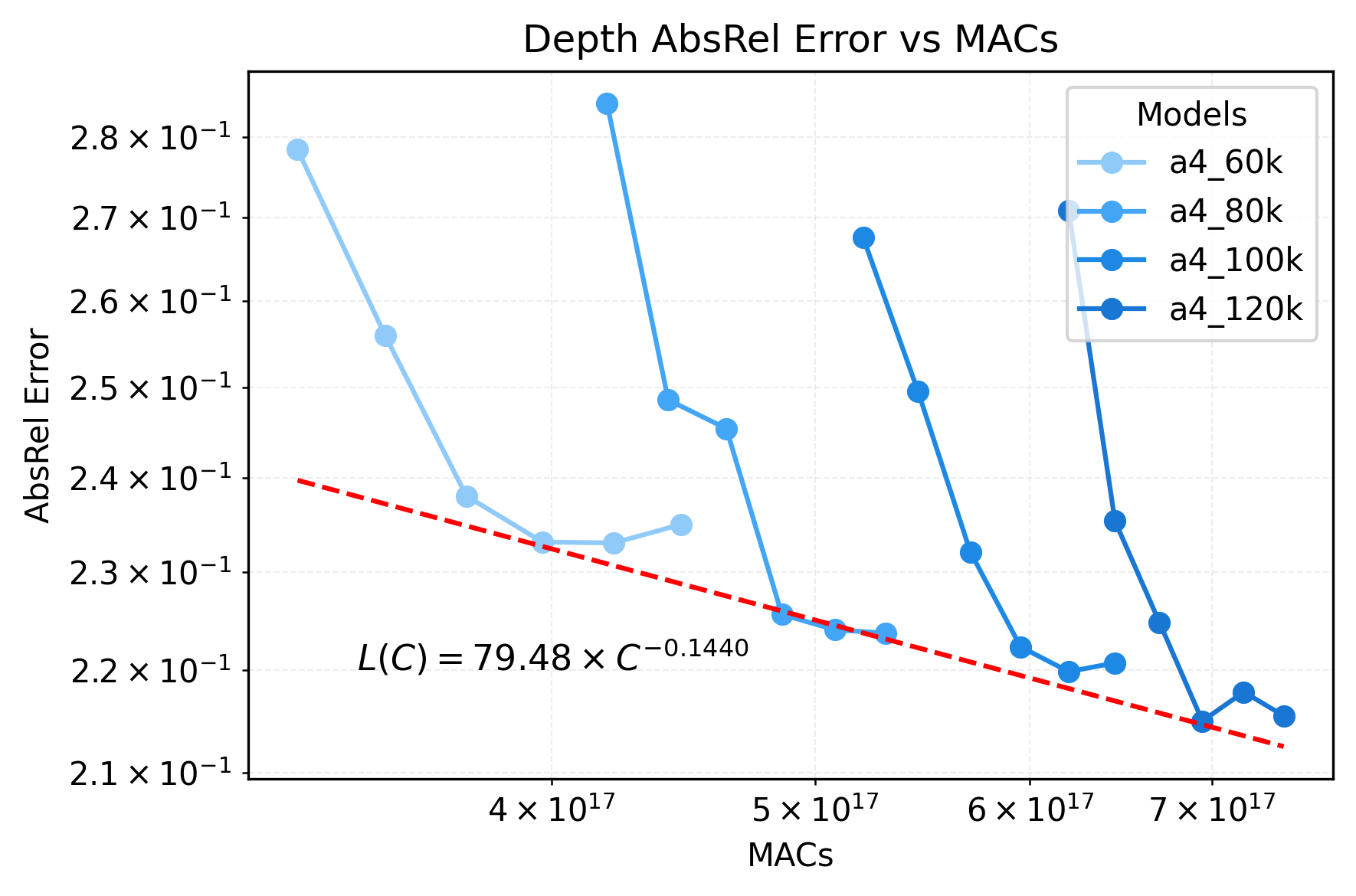}
        % \caption*{(a)}
    \end{minipage}
    \hfill
    \begin{minipage}[b]{0.48\textwidth}
        \centering
        \includegraphics[width=0.99\linewidth]{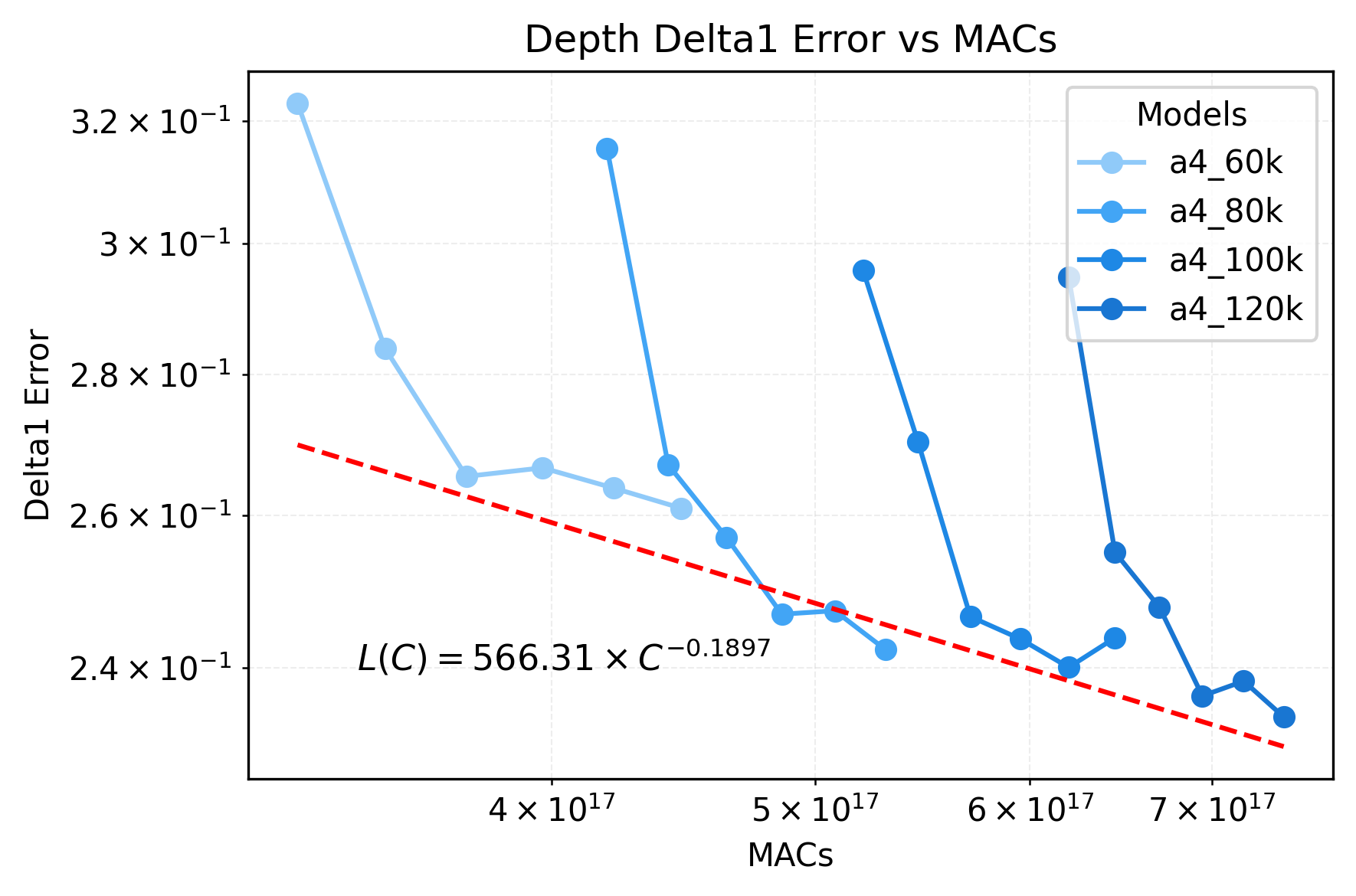}
        % \caption*{(b)}
    \end{minipage}
    \caption{\textbf{Effect of Scaling Model Pre-training Compute on Depth Estimation:} (a) Depth Absolute Relative Error vs. MACs. (b) Depth Delta1 Error vs. MACs. We pre-train four a4 models with 60K, 80K, 100K, and 120K steps. These models are then fine-tuned for 30K steps on the Hypersim depth estimation dataset. We observe a clear power law as we increase the DiT pre-training compute across depth estimation validation metrics.}
    \label{fig:dense_pretraining_steps}
\end{figure}

We also investigate the behavior of fine-tuning as we scale the number of pre-training steps for the DiT backbone. We train four models with the a4 configuration using a varied number of pre-training steps, keeping all other hyperparameters constant. We then fine-tune these four models on the same depth estimation dataset.Fig.~\ref{fig:dense_pretraining_steps} displays the power law scaling behavior of the validation metrics for depth estimation as we increase DiT pre-training steps. Our experiments show that having stronger pre-trained representations can be helpful when scaling fine-tuning compute.

\subsection{Effect of Image Resolution}
\begin{figure}[htbp]
    \centering
    \begin{minipage}[b]{0.48\textwidth}
        \centering
        \includegraphics[width=0.99\linewidth]{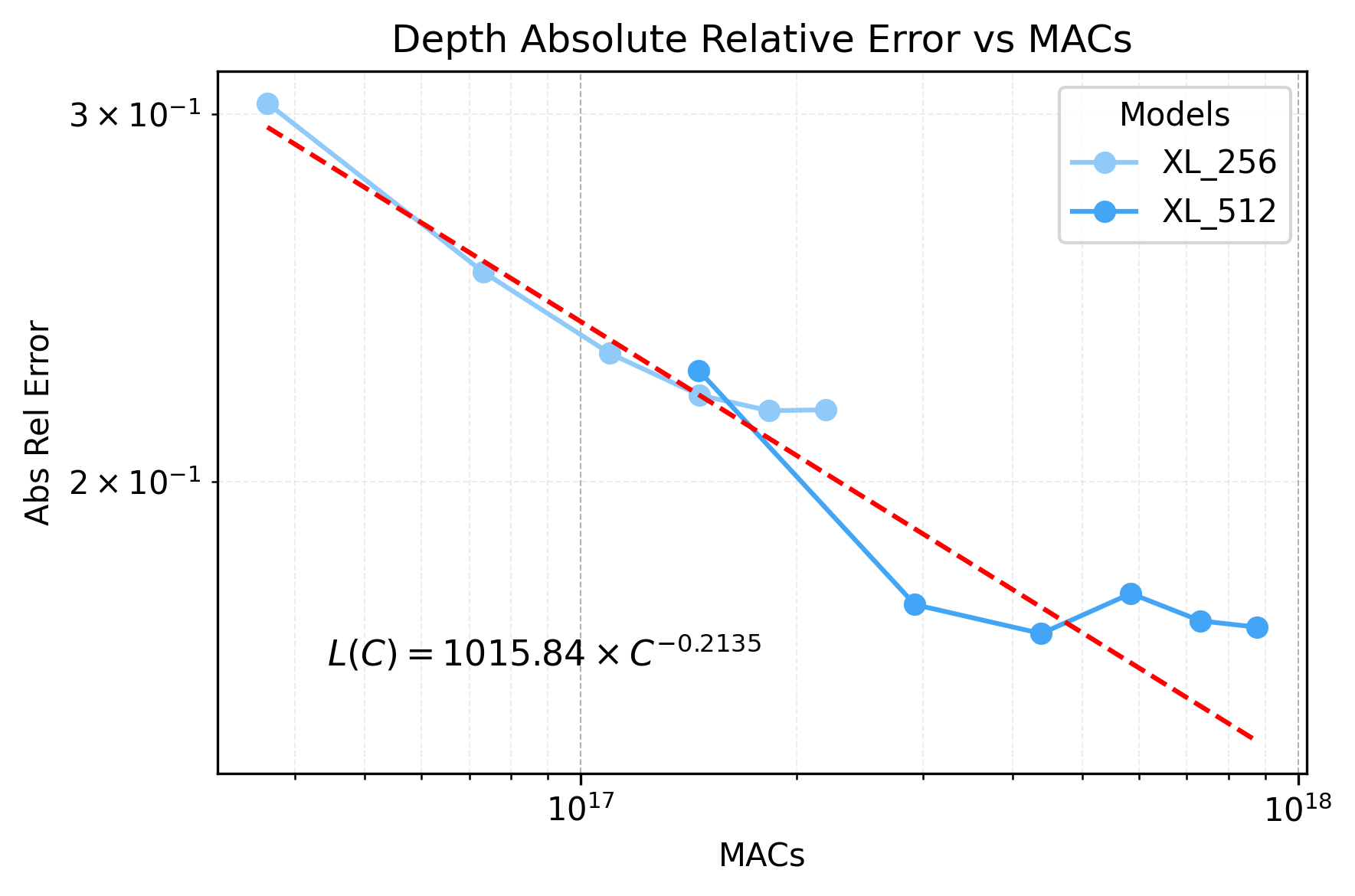}
        % \caption*{(a)}
    \end{minipage}
    \hfill
    \begin{minipage}[b]{0.48\textwidth}
        \centering
        \includegraphics[width=0.99\linewidth]{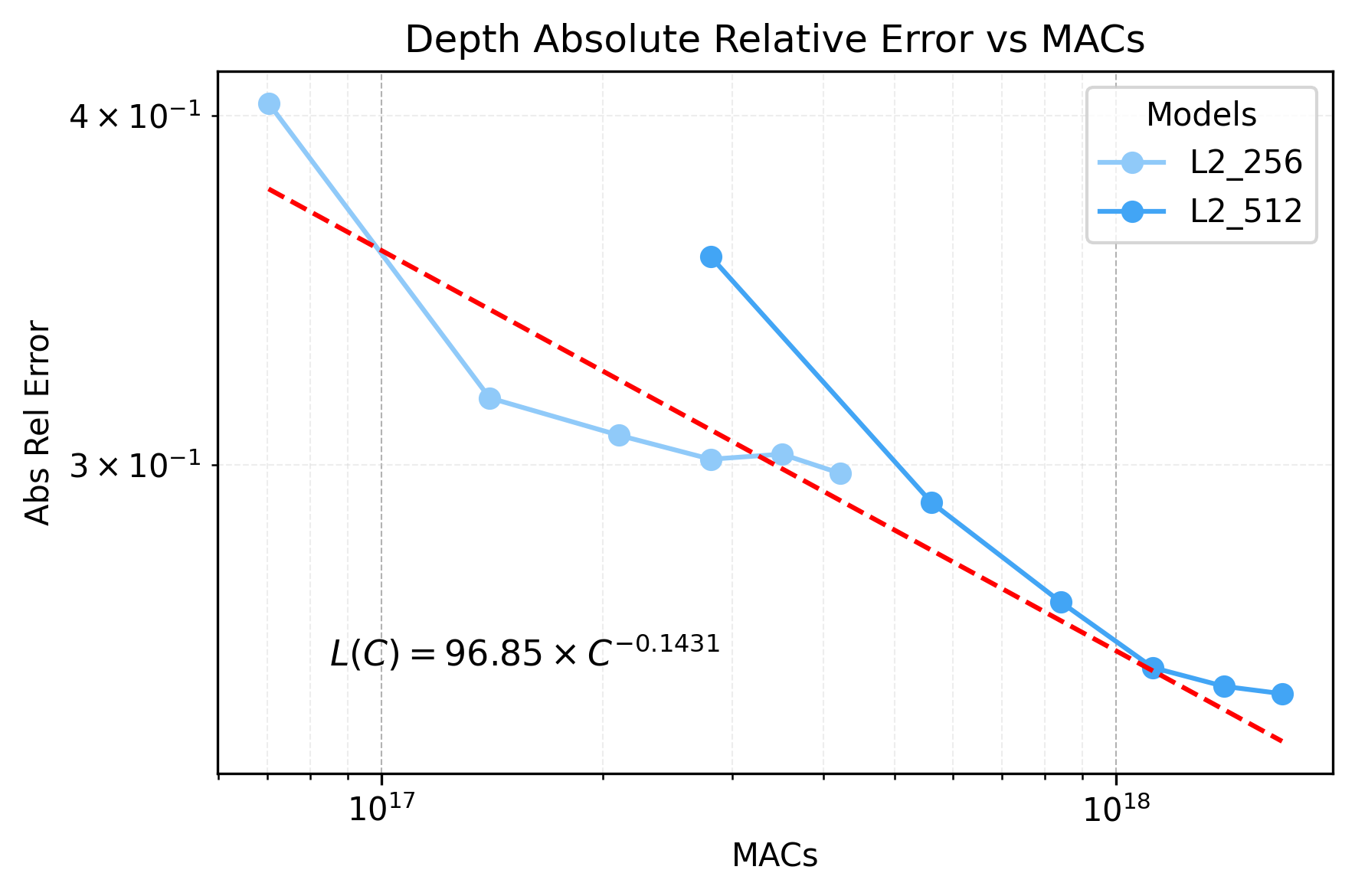}
        % \caption*{(b)}
    \end{minipage}
    \caption{\textbf{Effect of Image Resolution.} We fine-tune DiT-XL and DiT-MoE L/2 models with resolutions of $256 \times 256$ and $512 \times 512$. We observe a power law when increasing image resolution during training. By scaling the number of tokens per image by 4$\times$, we achieve strong performance on Depth Absolute Error, displaying the effect of increasing total dataset tokens for dense visual perception tasks such as depth estimation.}
    \label{fig:image_resolution}
\end{figure}

The sequence length of each image also affects the total compute spent during training. For each forward pass, we can scale the amount of compute used by simply increasing the resolution of the image, which will increase the number of tokens in the image embedding. By increasing the number of tokens, we can increase the amount of information the model can learn from at training time to build stronger internal representations, which can in turn improve downstream performance. We use dense DiT-XL models with resolutions of $256 \times 256$ and $512 \times 512$ from \citep{peebles2023scalable} and we pre-train DiT-MoE L/2-8E2A models with $256 \times 256$ and $512 \times 512$ resolutions following the recipe in \citep{FeiDiTMoE2024}. We then fine-tune each of these models with the corresponding resolution for the depth estimation task. Fig.~\ref{fig:image_resolution} displays that increasing image resolution to scale fine-tuning compute can provide significant gains on downstream depth estimation performance.

\subsection{Effect of Upcycling}
\begin{figure}[htbp]
    \centering
    \begin{minipage}[b]{0.48\textwidth}
        \centering
        \includegraphics[width=0.99\linewidth]{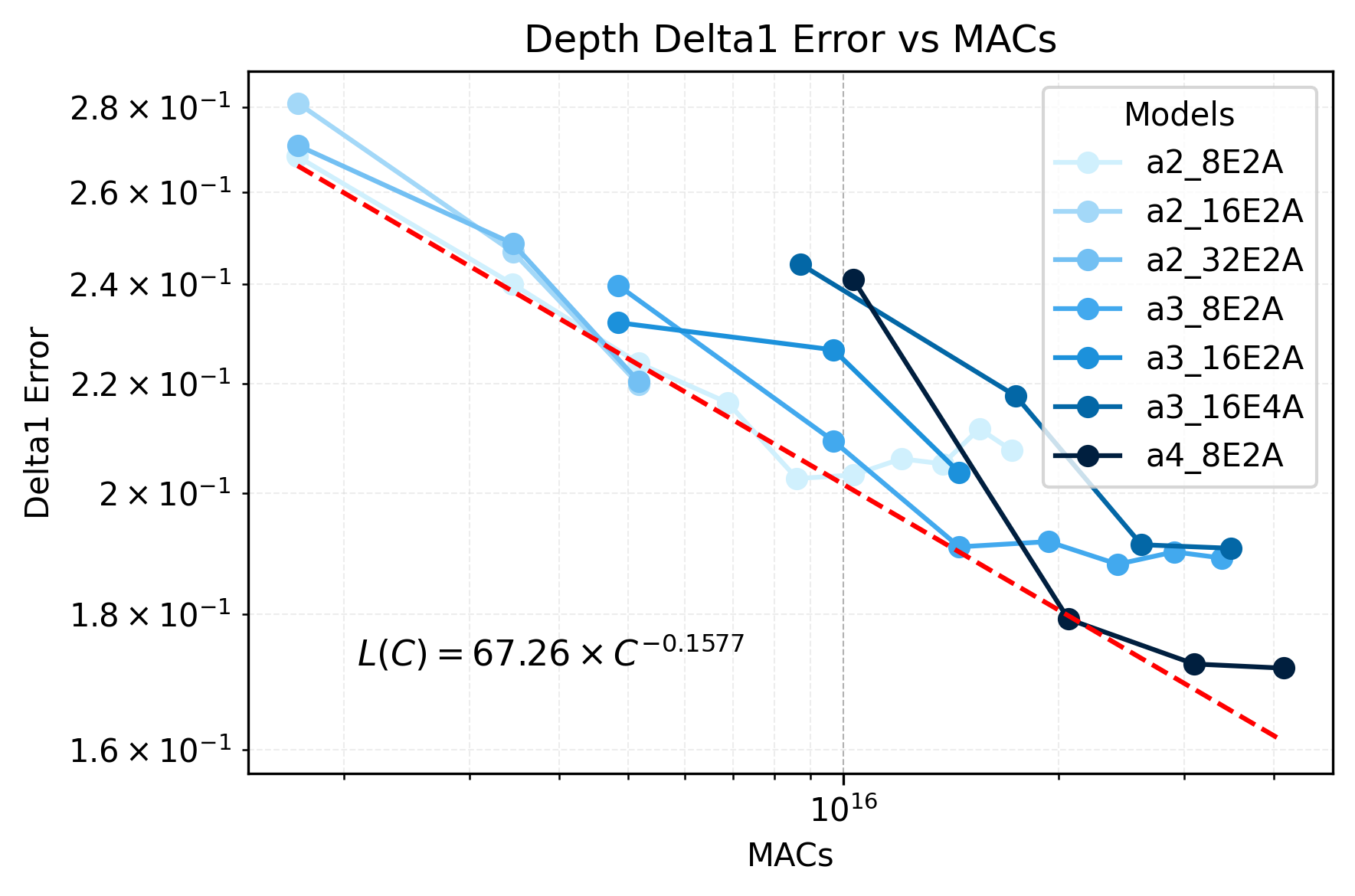}
        % \caption*{(a)}
    \end{minipage}
    \hfill
    \begin{minipage}[b]{0.48\textwidth}
        \centering
        \includegraphics[width=0.99\linewidth]{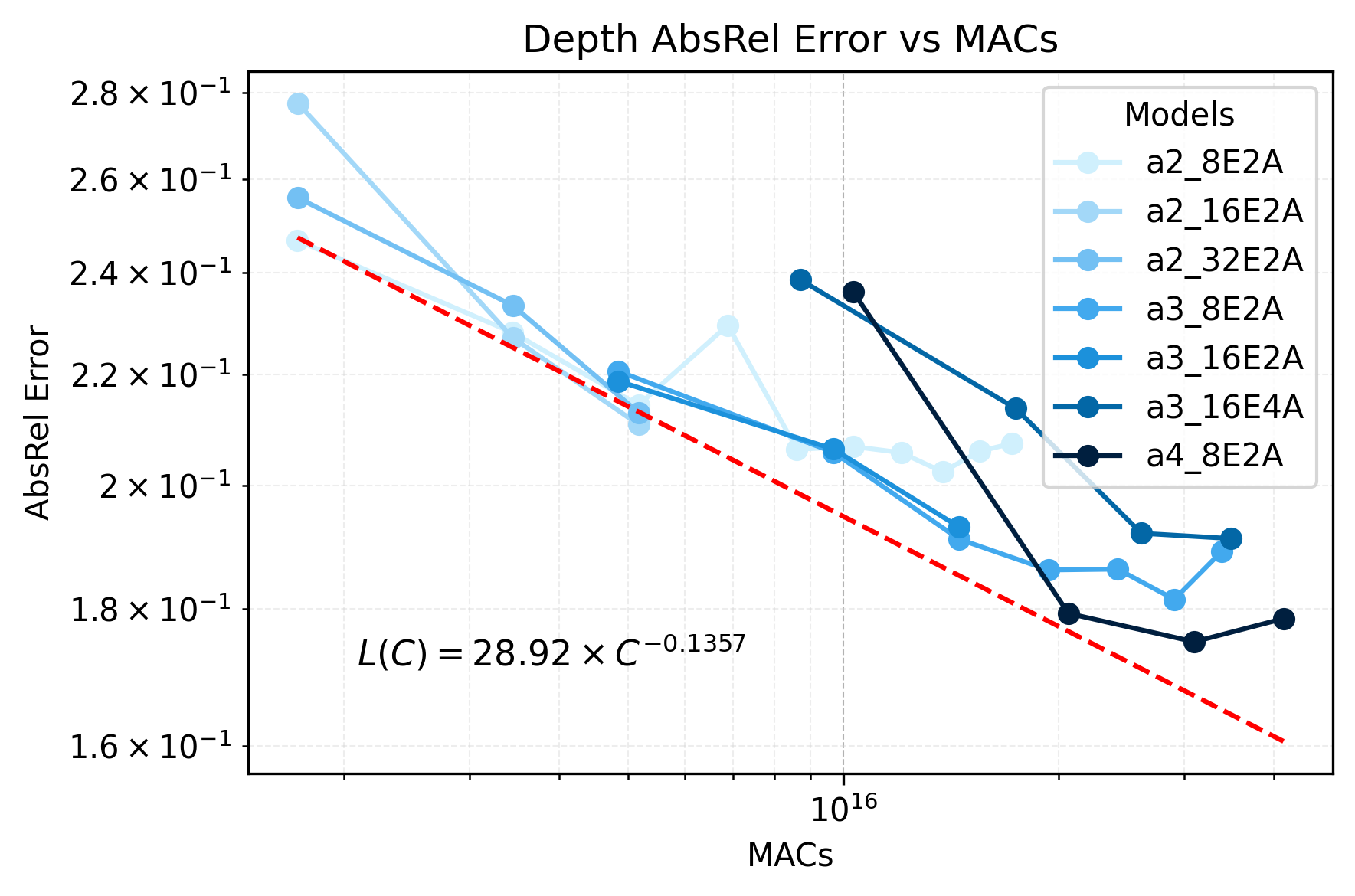}
        % \caption*{(b)}
    \end{minipage}
    \caption{\textbf{Effect of Upcycling.} We upcycle a2, a3, and a4 models fine-tuned for depth estimation with a varying number of total/active model experts. We continue fine-tuning each upcycled model for 15K iterations on the Hypersim depth estimation dataset. We observe a clear scaling law in the validation metrics as we increase fine-tuning compute with upcycling. The upcycled models can also achieve equivalent or superior performance to our dense a5 and a6 checkpoints, each of which utilize more compute during pre-training and fine-tuning. Increasing the total model experts and total active experts can also improve the downstream performance.}
    \label{fig:upcycle_fine_tune}
\end{figure}

Sparse MoE models are efficient options for increasing the capacity of a model, but pre-training an MoE model from scratch can be expensive. One way to alleviate this issue is Sparse MoE Upcycling \citep{komatsuzaki2023sparseupcyclingtrainingmixtureofexperts}. Upcycling converts a dense transformer checkpoint to an MoE model by copying the MLP layer in each transformer block $E$ times, where $E$ is the number of experts, and adding a learnable router module that sends each token to the top-$k$ selected experts. The outputs of the selected experts are then combined in a weighted sum at the end of each MoE block. We upcycle various dense DiT models after they are fine-tuned for depth estimation and then continue fine-tuning the upcycled model. Fig.~\ref{fig:upcycle_fine_tune} displays the scaling laws for upcycling, providing an average improvement of~5.3\% on Absolute Relative Error and~8.6\% on Delta1 error.

\section{Scaling Test-Time Compute}

\begin{figure}[!htbp]
    \centering
    \includegraphics[width=0.9\linewidth]{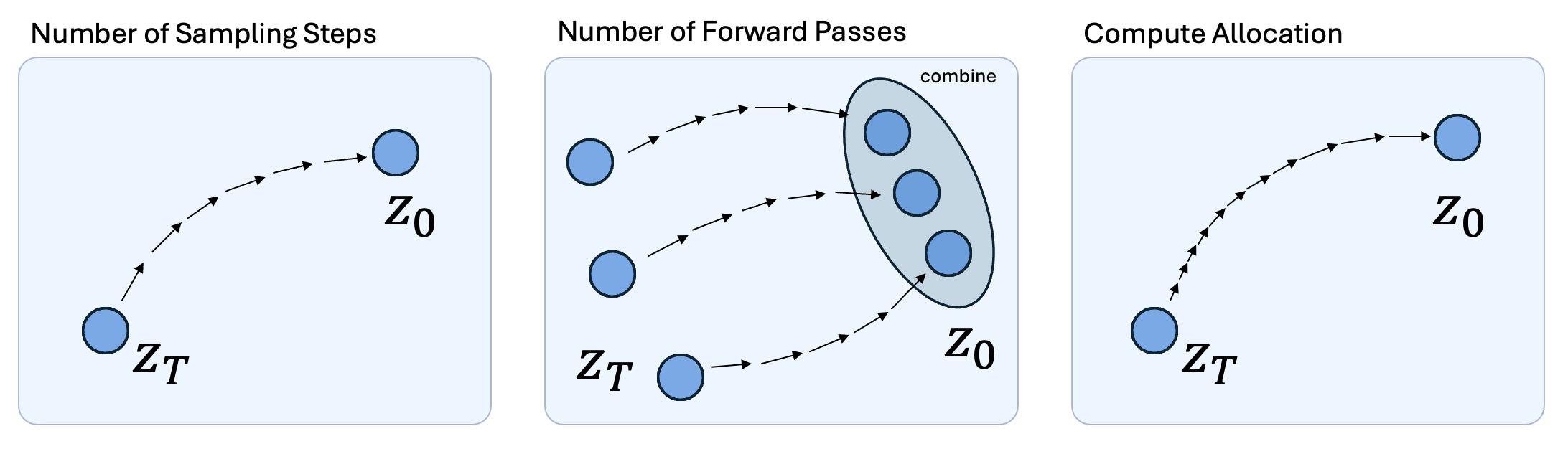}
    % \caption*{(a)}
    \caption{\textbf{Inference Scaling:} Diffusion models by design allow efficient scaling of test-time compute. First, we can simply increase the number of denoising steps to increase the compute spent at inference. Since we are estimating deterministic outputs, we can then initialize multiple noise latents and ensemble the predictions to get a better estimation. Finally, we can also reallocate our test-time compute budget for low and high frequency denoising by modifying the noise variance schedule.
    }
    \label{fig:inference_pipeline}
\end{figure}

Scaling test-time compute has been explored for autoregressive Large Language Models (LLMs) to improve performance on long-horizon reasoning tasks \citep{brown2024largelanguagemonkeysscaling, snell2024scalingllmtesttimecompute, elswag, openai2024o1}. In this section, we show how to reliably improve diffusion model performance for perceptual tasks by scaling test-time compute. We summarize our approach in Fig.~\ref{fig:inference_pipeline}. We use the Stable-Diffusion VAE to encode the input image into latent space \citep{rombach2022high}. Then, we sample a target noise latent from a standard Gaussian distribution, which is iteratively denoised with DDIM~\citep{song2021denoising} to generate the downstream prediction. 

\subsection{Effect of Scaling Inference Steps}

\begin{figure}[htbp]
    \centering
    \begin{minipage}[b]{0.48\textwidth}
        \centering
        \includegraphics[width=0.9\linewidth]{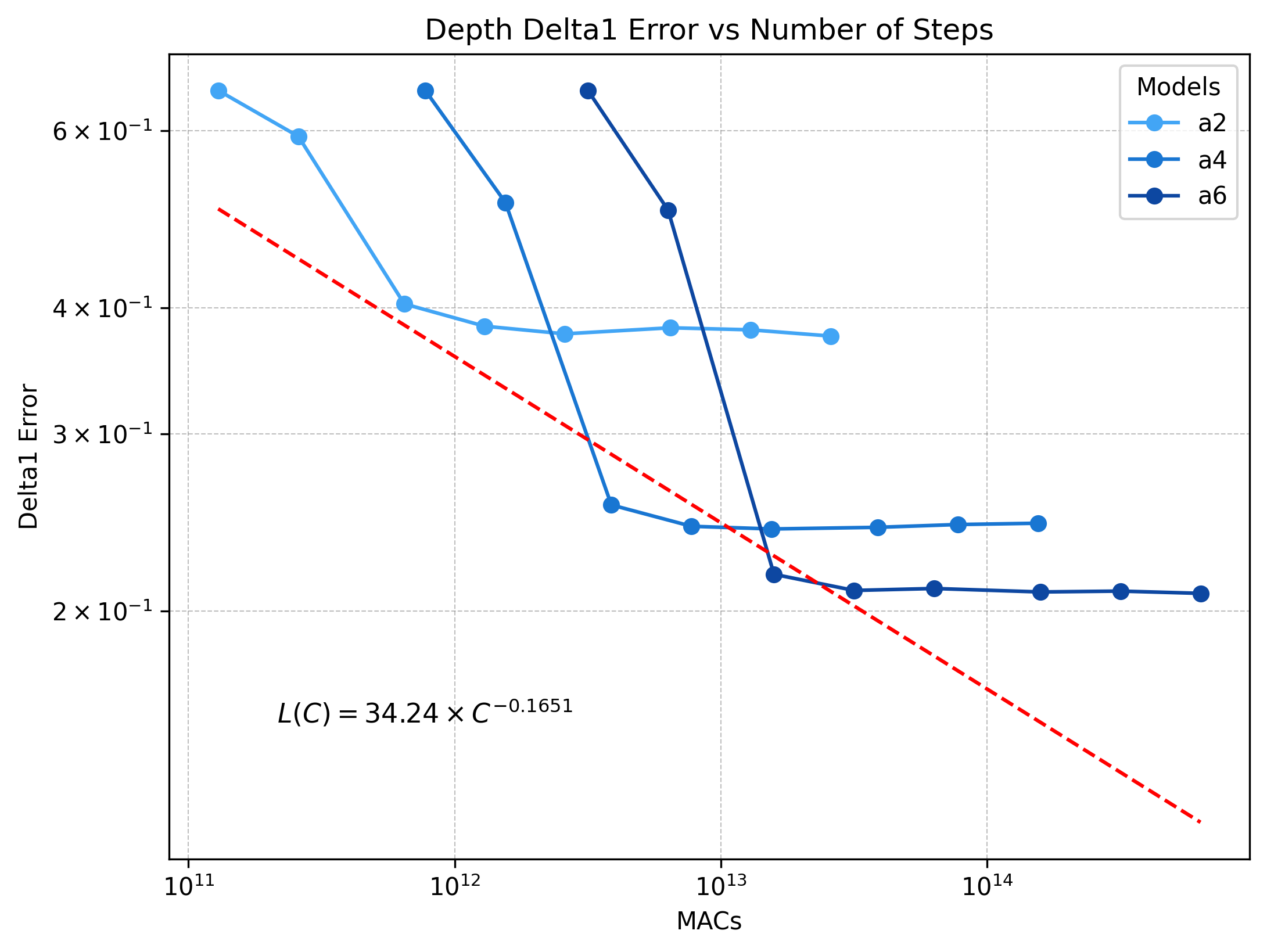}
    \end{minipage}
    \hfill
    \begin{minipage}[b]{0.48\textwidth}
        \centering
        \includegraphics[width=0.9\linewidth]{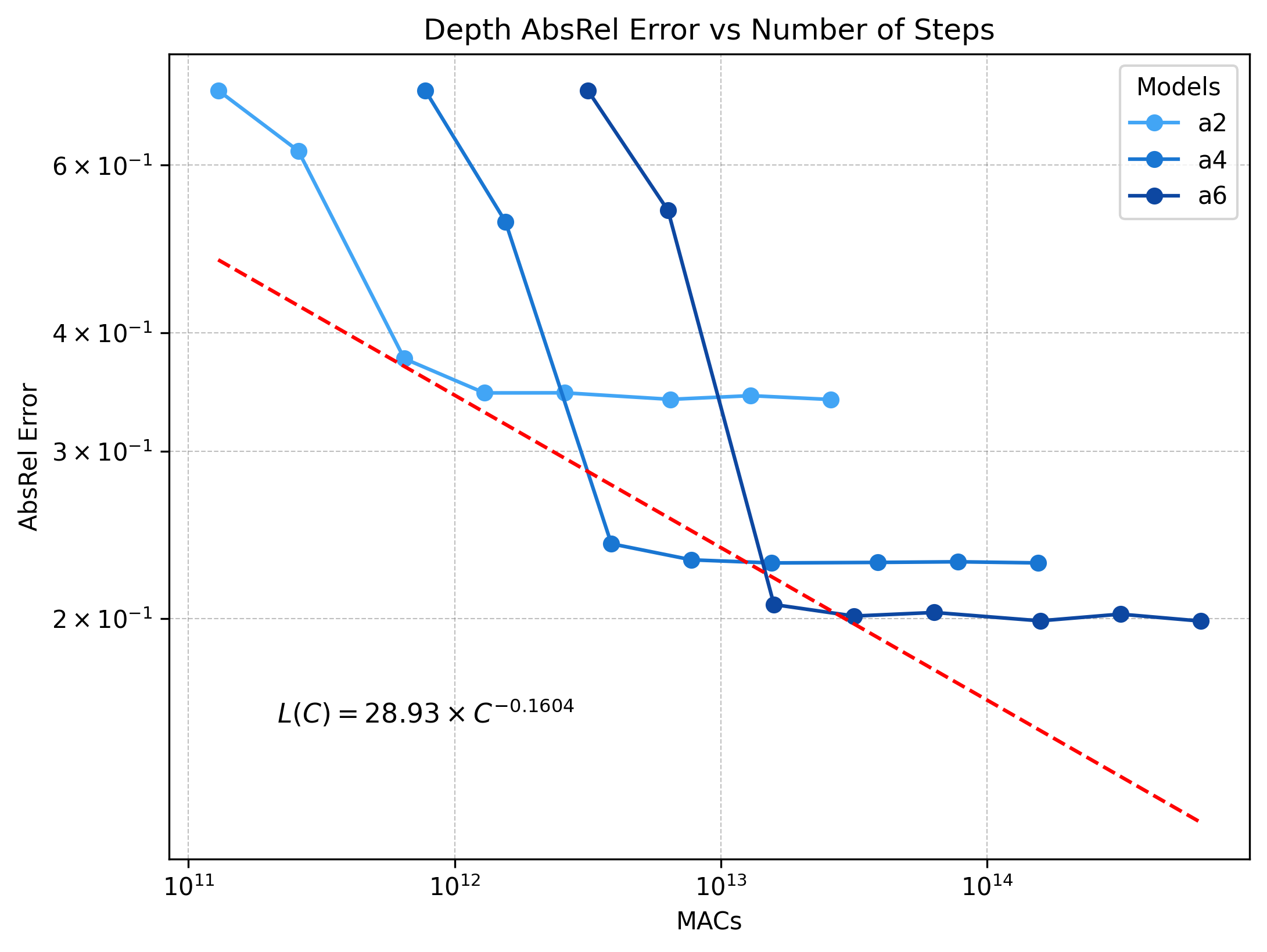}
    \end{minipage}
    \caption{\textbf{Effect of Number of Sampling Steps.} (a) Delta1 Error vs. Number of Steps. (b) Absolute Relative Error vs. Number of Steps. For each model, we sample for $T \in \left[1, 2, 5, 10, 20, 50, 100\right]$ steps with the DDIM sampler. We show a clear power law scaling behavior in (a) and (b), displaying the effectiveness of scaling test-time compute by increasing the number of diffusion sampling steps.}
    \label{fig:sampling_steps}
\end{figure}

The most natural way of scaling diffusion inference is by increasing denoising steps.
Since the model is trained to denoise the input at various timesteps, we can scale the number of diffusion denoising steps at test-time to produce finer, more accurate predictions. This coarse-to-fine denoising paradigm is also reflected in the generative case, and we can take advantage of it for the discriminative case by increasing the number of denoising steps. In Fig.~\ref{fig:sampling_steps}, we observe that increasing the total test-time compute by simply increasing the number of diffusion sampling steps provides substantial gains in depth estimation performance.

\subsection{Effect of Test Time Ensembling}

We also explore scaling inference compute with test-time ensembling. We exploit the fact that denoising different noise latents will generate different downstream predictions. In test-time ensembling, we compute $N$ forward passes for each input sample and reduce the outputs through one of two methods. The first technique is naive ensembling where we use the pixel-wise median across all outputs as the prediction. The second technique presented in Marigold \citep{ke2024repurposing} is median compilation, where we collect predictions $\displaystyle \{\hat{\vd_1}, \dots , \hat{\vd_N}\}$ that are affine-invariant, jointly estimate scale and shift parameters $\hat{s_i}$ and $\hat{t_i}$, and minimize the distances between each pair of scaled and shifted predictions $\displaystyle (\hat{\vd'_i}, \hat{\vd'_j})$ where $\hat{\vd'} = \hat{\vd} \times \hat{s} + \hat{t}$. For each optimization step, we take the pixel-wise median $\vm(x,y) = \text{median}(\hat{\vd'_1(x, y)}, \dots, \hat{\vd'_N(x, y)})$ to compute the merged depth $\vm$. 
Since it requires no ground truth, we scale ensembling by increasing $N$ to utilize more test-time compute.

\begin{figure}[htbp]
    \centering
    \begin{minipage}[b]{0.48\textwidth}
        \centering
        \includegraphics[width=0.9\linewidth]{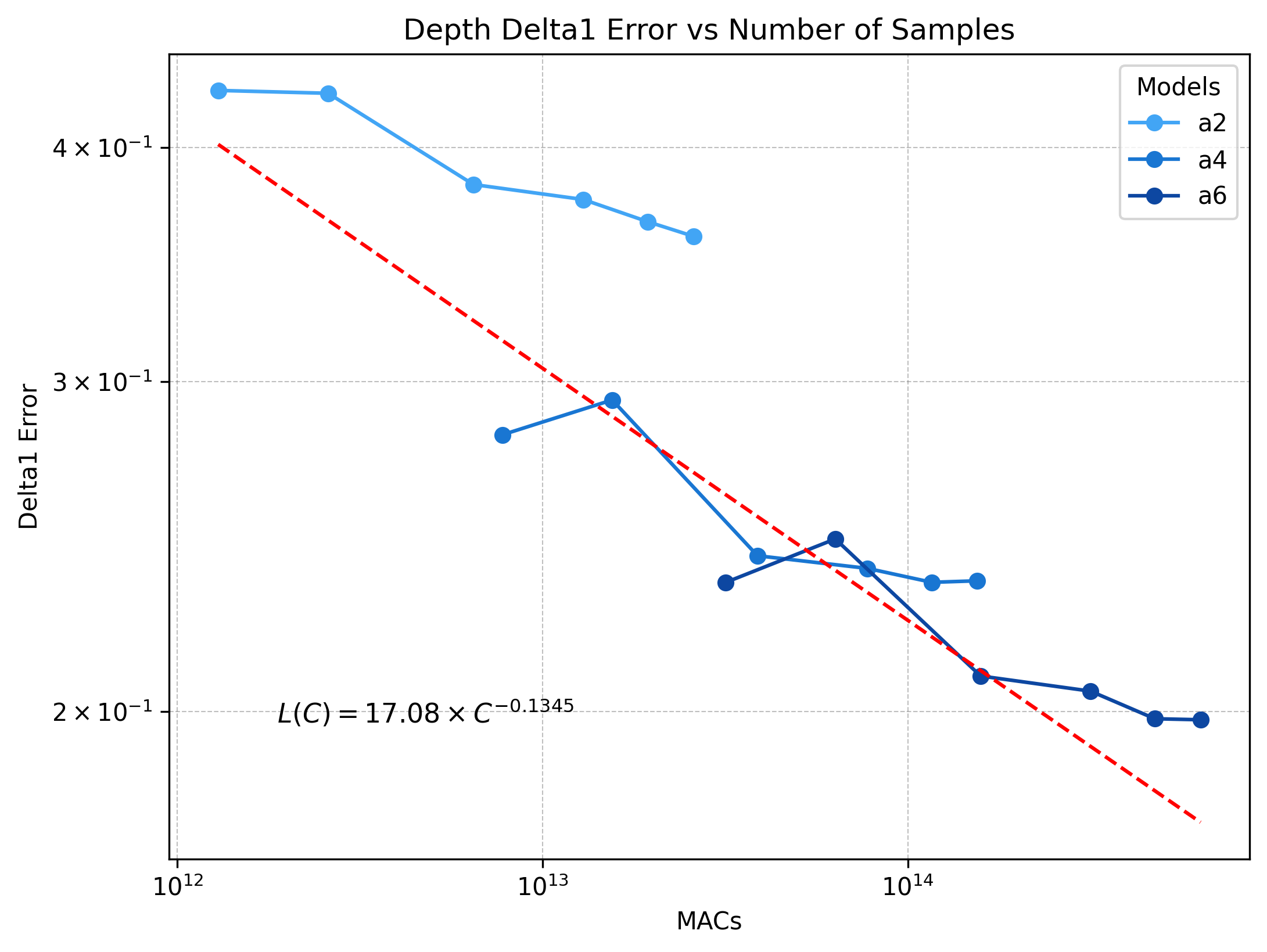}
        % \caption*{(a)}
    \end{minipage}
    \hfill
    \begin{minipage}[b]{0.48\textwidth}
        \centering
        \includegraphics[width=0.9\linewidth]{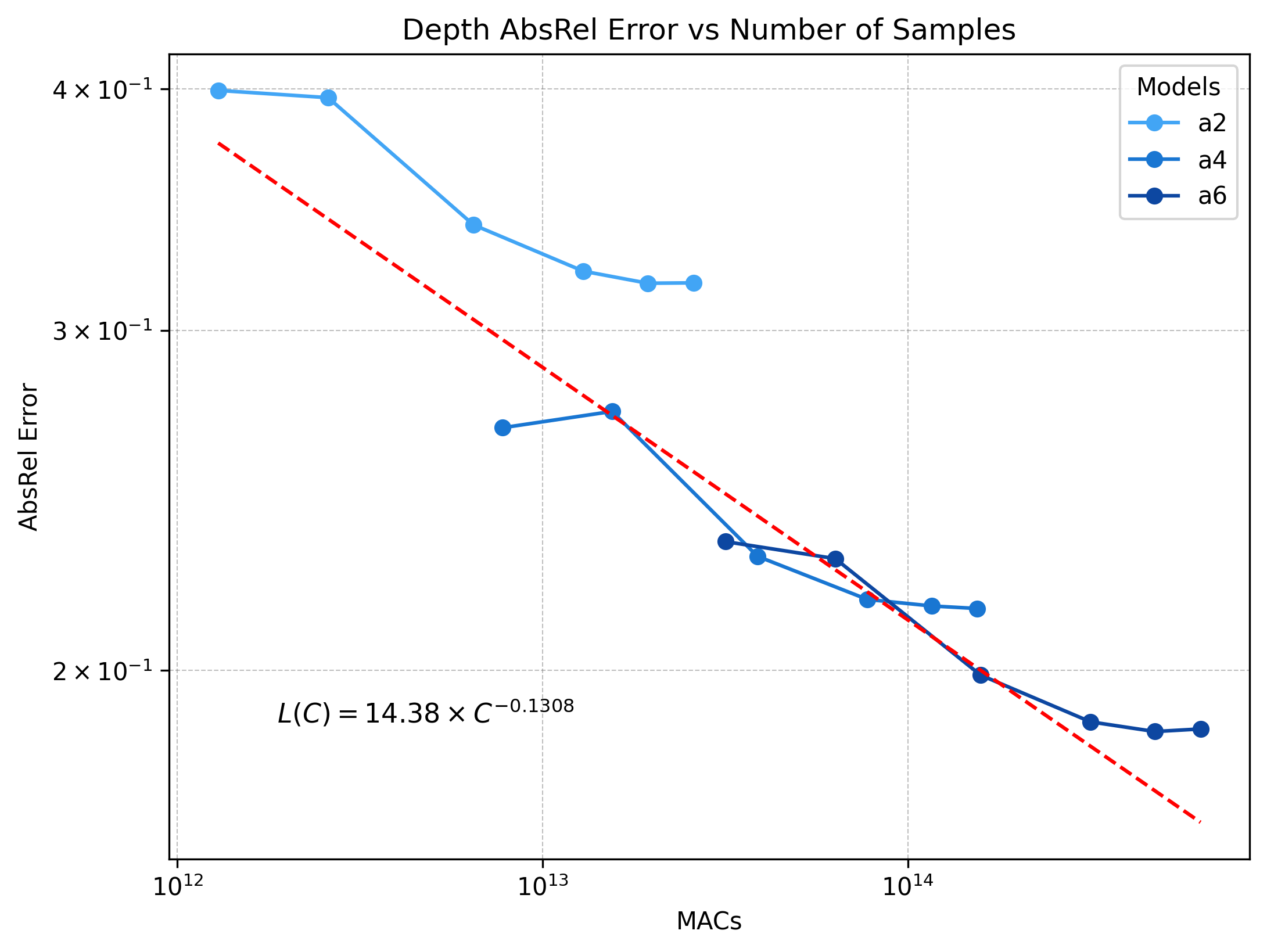}
        % \caption*{(b)}
    \end{minipage}
    \caption{\textbf{Effect of Test Time Ensembling.} (a) Delta1 Error vs. Number of Forward Passes. (b) Absolute Relative Error vs. Number of Forward Passes. Ensembling multiple predictions from distinct noise initializations displays power law scaling behavior. We apply test-time ensembling values of $N \in \left[ 1, 2, 5, 10, 15, 20 \right]$. } % mistake in graph title for abs rel
    \label{fig:tt_ensembling}
\end{figure}

\subsection{Effect of Noise Variance Schedule}
\begin{figure}[htbp]
    \centering
    \begin{minipage}[b]{0.48\textwidth}
        \centering
        \includegraphics[width=0.99\linewidth]{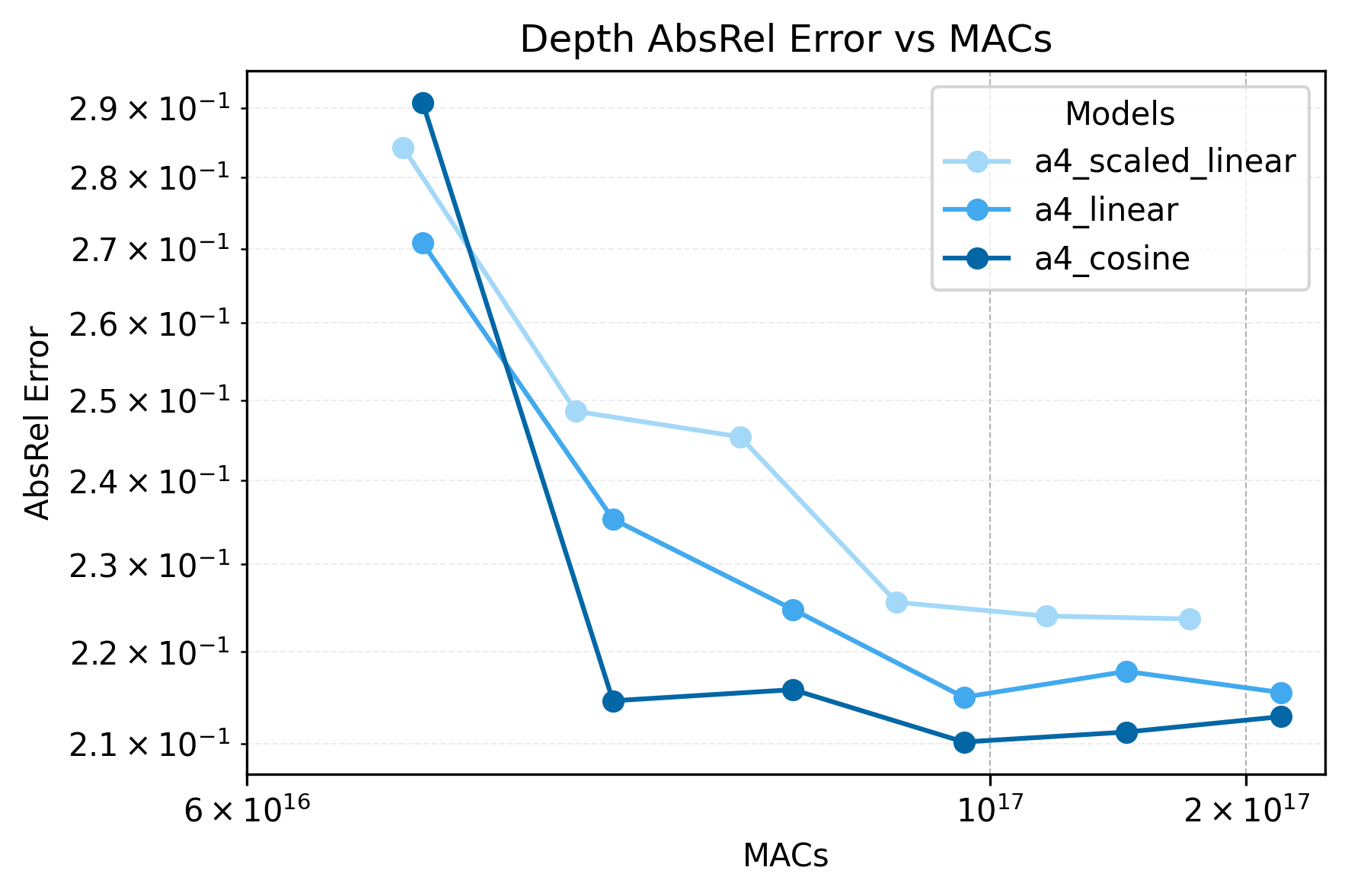}
        % \caption*{(a)}
    \end{minipage}
    \hfill
    \begin{minipage}[b]{0.48\textwidth}
        \centering
        \includegraphics[width=0.99\linewidth]{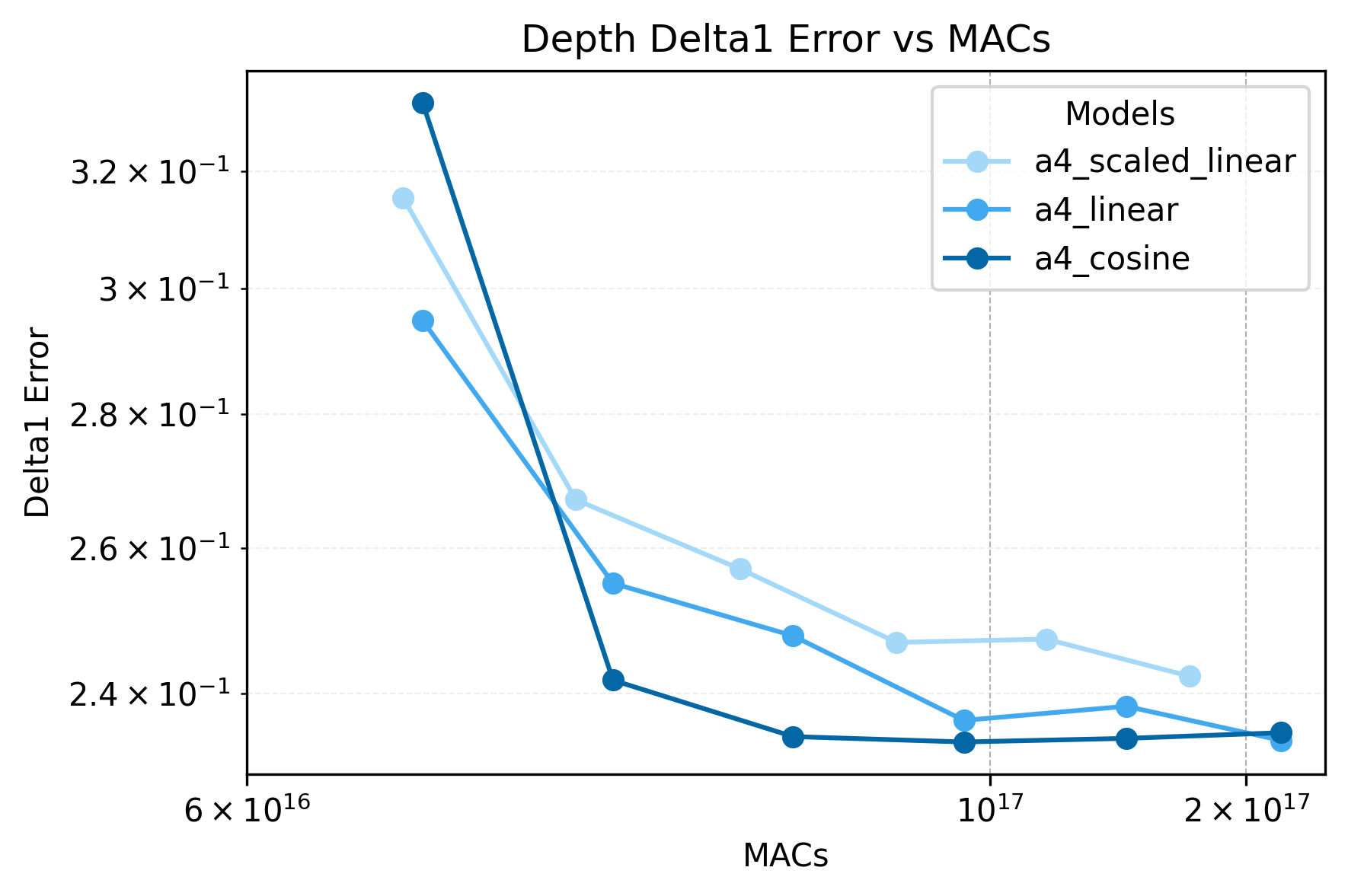}
        % \caption*{(b)}
    \end{minipage}
    \caption{\textbf{Effect of Noise Variance (Beta) Schedule.} We fine-tune a4 models with three different beta schedules: linear, scaled linear, cosine. Reallocating compute with the cosine schedule to spend more time denoising at earlier timesteps significantly improved Delta1 and Absolute Relative Error rates.}
    \label{fig:noise_variance}
\end{figure}

We can also scale test-time compute by increasing compute usage at different points of the denoising process. In diffusion noise schedulers, we can define a schedule for the variance of the Gaussian noise applied to the image over the total diffusion timesteps $T$. Tuning the noise variance schedule allows for reorganizing compute by allocating more compute to denoising steps earlier or later in the noise schedule. We experiment with three different noise level settings for DDIM: linear, scaled linear, and cosine. Cosine scheduling from \citep{nichol2021improveddenoisingdiffusionprobabilistic} linearly declines from the middle of the corruption process, ensuring the image is not corrupted too quickly as in linear schedules. Fig.~\ref{fig:noise_variance} shows that the cosine noise variance schedule outperforms linear schedules for DDIM on the depth estimation task under a fixed compute budget.

\section{Putting It All Together}

Using the lessons from our scaling experiments on depth estimation, we train diffusion models for optical flow prediction and amodal segmentation. We show that using diffusion models while considering efficient methods to scale training and test-time compute can provide substantial performance gains on visual perception tasks, achieving improved or similar performance as current state-of-the-art techniques. Our experiments provide insight on how to efficiently apply diffusion models for these visual perception tasks under limited compute budgets. Finally, we train a unified expert model, capable of performing all three visual perception tasks previously mentioned, displaying the generalizability of our method. Our results prove the effectiveness of our training and test-time scaling strategies, removing the need to use pre-trained models trained on internet-scale datasets to enable high-quality visual perception in diffusion models. Fig.~\ref{fig:generated_samples} displays the predicted samples from our models.

\subsection{Depth Estimation}
We combine our findings from the ablation studies on depth estimation to create a model with the best training and inference configurations. We train a DiT-XL model from \citep{peebles2023scalable} on depth estimation data from Hypersim for 30K steps with a batch size of 1024, resolution of $512 \times 512$, and a learning rate exponentially decaying from $1.2e$-$4$ to $1.2e$-$6$. We use median compilation ensembling with a cosine noise variance schedule. From our scaling experiments, we found the optimal configuration for inference to be 200 denoising steps with $N=5$ samples for ensembling. As shown in Table~\ref{table:depth_numbers}, our model achieves the same validation performance as Marigold on the Hypersim dataset and better performance on the ETH3D test set while being trained with lower resolution images and approximately \textbf{three orders of magnitude less pre-training data and compute}.

\begin{table*}[h]
\centering
\small % Reduces the font size for the table
\setlength{\tabcolsep}{5pt} % Adjusts the column separation for tighter spacing
\setlength{\tabcolsep}{3pt}
\begin{tabular}{lcccccccccc}
\toprule
\textbf{Method} & \multicolumn{2}{c}{\textbf{Hypersim}} & \multicolumn{2}{c}{\textbf{ETH3D}} & \multicolumn{2}{c}{\textbf{NYUv2}} & \multicolumn{2}{c}{\textbf{ScanNet}} & \multicolumn{2}{c}{\textbf{DIODE}} \\
 & AbsRel $\downarrow$ & $\delta1\uparrow$ & AbsRel $\downarrow$ & $\delta1\uparrow$ & AbsRel $\downarrow$ & $\delta1\uparrow$ & AbsRel $\downarrow$ & $\delta1\uparrow$ & AbsRel $\downarrow$ & $\delta1\uparrow$ \\
\midrule
DiverseDepth & $-$ & $-$ & 22.8 & 69.4 & 11.7 & 87.5 & 10.9 & 88.2 & 37.6 & 63.1 \\
MiDaS & $-$ & $-$ & 18.4 & 75.2 & 11.1 & 88.5 & 12.1 & 84.6 & 33.2 & 71.5 \\
LeReS & $-$ & $-$ & 17.1 & 77.7 & 9.0 & 91.6 & 9.1 & 91.7 & 27.1 & 76.6 \\
Omnidata & $-$ & $-$ & 16.6 & 77.8 & 7.4 & 94.5 & 7.5 & 93.6 & 33.9 & 74.2 \\
HDN & $-$ & $-$ & 12.1 & 83.3 & 6.9 & 94.8 & 8.0 & 93.9 & 24.6 & 78.0 \\
DPT & $-$ & $-$ & 7.8 & 94.6 & 9.8 & 90.3 & 8.2 & 93.4 & 18.2 & 75.8 \\
Marigold & 13.5 & 87.5 & 6.5 & 96.0 & 5.5 & 96.4 & 6.4 & 95.1 & 30.8 & 77.3 \\ \hline
\textbf{Ours} & 13.6 & 87.6 & 4.8 & 97.8 & 6.8 & 95.0 & 7.7 & 93.7 & 31.0 & 77.2 \\
\bottomrule
\end{tabular}
\caption{\textbf{Depth Estimation Performance Comparison on Multiple Datasets.} We achieve state-of-the-art performance on the ETH3D dataset and competitive performance across all other benchmarks. Notably, we closely match the performance of Marigold across all datasets with significantly less training compute.}
\label{table:depth_numbers}
\end{table*}

\subsection{Optical Flow Prediction}
Optical flow estimation involves predicting the motion of objects between consecutive frames in a video, represented as a dense vector field indicating pixel-wise displacement. We use a similar configuration as the depth estimation model for optical flow training. We train a DiT-XL model on the FlyingChairs dataset for 40K steps with batch size of 1024, resolution of $512 \times 512$, and learning rate exponentially decaying from $1.2e$-$4$ to $1.2e$-$6$. We compare our model's performance with other specialized optical flow prediction techniques in Table~\ref{table:optical_flow}.

\begin{table}[!htb]
\centering
\footnotesize
\setlength{\tabcolsep}{8pt}

\begin{minipage}{0.5\textwidth} % Adjust the width to control alignment
\centering
\setlength{\tabcolsep}{3pt}
\begin{tabular}{l c} 
\toprule[0.4mm]
\textbf{Method} & \textbf{FlyingChairs EPE} $\downarrow$ \\ \midrule
DeepFlow & 3.53 \\
FlowNetS & 2.71 \\
FlowNetS+v & 2.86 \\
FlowNetS+ft & 3.04 \\
FlowNetS+ft+v & 3.03 \\
FlowNetC & 2.19 \\
FlowNetC+v & 2.61 \\
FlowNetC+ft & 2.27 \\
FlowNetC+ft+v & 2.67 \\ \hline
\textbf{Ours (w/o ensembling)} & 3.45 \\
\textbf{Ours (w/ ensembling)} & 3.08 \\
\bottomrule[0.4mm]
\end{tabular}
\end{minipage}%
\begin{minipage}{0.45\textwidth} % Adjust width to match the table width
\vspace{0.5em} % Add space to align with the table vertically if needed
\caption{\textbf{Optical Flow Comparison with Specialized Techniques.} We evaluate our optical flow model on the FlyingChairs validation set. Our model achieves similar end-point error as specialized methods, including DeepFlow \citep{Weinzaepfel_2013_ICCV} and FlowNet \citep{fischer2015flownetlearningopticalflow}. We train with significantly less data compared to other specialized methods, which use a several optical flow datasets. We generate predictions with and without test-time ensembling.}
\label{table:optical_flow}
\end{minipage}

\end{table}

\subsection{Amodal Segmentation}
Amodal segmentation is the process of predicting the complete shape and extent of objects in an image, including the portions that are occluded or not directly visible, which can require higher-level reasoning for complex scenes. We fine-tune a DiT-XL model on the pix2gestalt dataset \citep{ozguroglu2024pix2gestalt} for 6K steps with a batch size of 4096, resolution of $256 \times 256$, and learning rate exponentially decaying from $1.2e$-$4$ to $1.2e$-$6$. We compare our model with other methods in Table~\ref{table:amodal_segmentation}.

\begin{table}[!htb]
\centering
\small % Reduces the font size for the table
\setlength{\tabcolsep}{5pt} % Adjusts the column separation for tighter spacing

\begin{minipage}{0.55\textwidth}
\centering
\begin{tabular}{lccc}
\toprule
\textbf{Method} & \textbf{COCO-A} & \textbf{P2G} & \textbf{MP3D} \\
\midrule
PCNet & 81.35 & $-$ & $-$ \\
PCNet-Sup & 82.53 & $-$ & $-$ \\
SAM & 67.21 & $-$ & $-$ \\
SD-XL Inpainting & 76.52 & $-$ & $-$ \\
pix2gestalt & 82.9 & 88.7 & 61.5 \\ \hline
\textbf{Ours} & 82.9 & 88.6 & 63.9 \\
\bottomrule
\end{tabular}
\end{minipage}%
\hfill
\begin{minipage}{0.45\textwidth}
\caption{\textbf{Amodal Segmentation Performance (mIOU) Comparison Across Different Datasets.} This table compares mIOU performance across COCO-A, Pix2Gestalt, and MP3D datasets, showing the effectiveness of various methods. Our method is able to achieve competitive performance across all tasks, while training only on Pix2Gestalt.}
\label{table:amodal_segmentation}
\end{minipage}
\end{table}

\subsection{One Model for All}
\begin{figure}[htbp]
    % Single combined image
    \centering
    \includegraphics[width=0.9\textwidth]{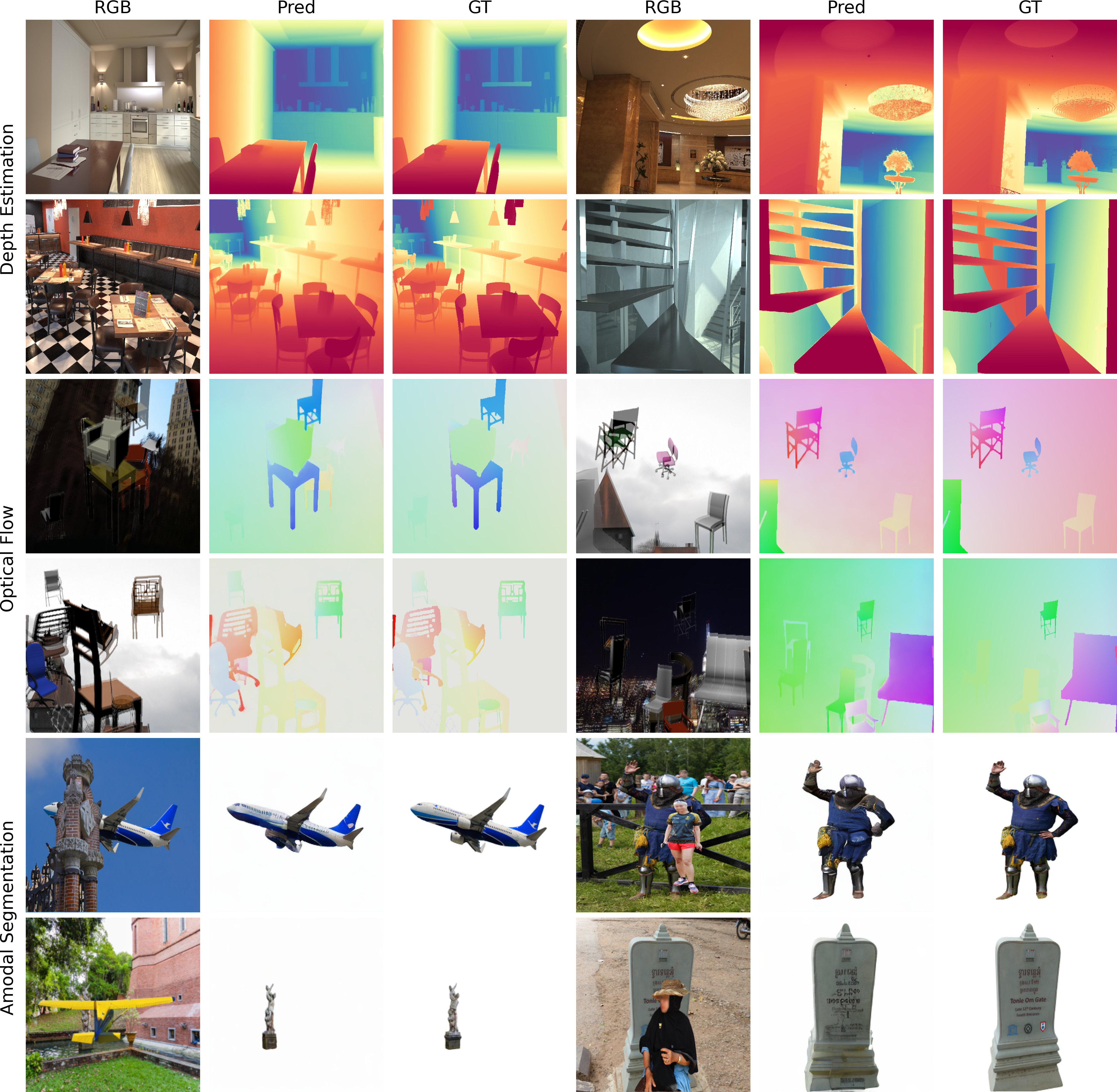}
    \caption{\textbf{Depth Estimation, Optical Flow Estimation, and Amodal Segmentation Examples:} Each row showcases results from our models for different tasks. (a) Depth estimation, with relative scale and shift. (b) Optical flow, with scale and shift. (c) Amodal segmentation, where the model sees an RGB image and segmentation of the occluded object; the task is to predict the amodal image.}
    
    \label{fig:generated_samples}
\end{figure}

We train a unified DiT-XL model for each of the different tasks. We train this model on a mixed dataset consisting of all three tasks. To train this generalist model, we modify the DiT-XL architecture by replacing the patch embedding layer with a separate \verb|PatchEmbedRouter| module, which routes each VAE embedding to a specific input convolutional layer based perception task. This ensures the DiT-XL model is able to distinguish between the task-specific embeddings during fine-tuning. We use a similar training recipe as the previous experiments, using images with $512 \times 512$ resolution and a learning rate exponentially decaying from $1.2e$-$4$ to $1.2e$-$6$. Then, we upcycle the fine-tuned DiT-XL checkpoint to an DiT-XL-8E2A model, and continue fine-tuning for another 4K iterations. We display the generated predictions in Fig.~\ref{fig:generated_samples} which exemplify the generalizability and transferability of our scaling techniques across a variety of perception tasks.

\vspace{-0.3cm}
\section{Conclusion}
\vspace{-0.2cm}
In our work, we examine the scaling properties of diffusion models for visual perception tasks. We explore various approaches to scale diffusion training, including increasing model size, mixture-of-experts models, increasing image resolution, and upcycling. We also efficiently scale test-time compute by exploiting the iterative nature of diffusion, which significantly improves downstream performance. Our experiments provide strong evidence of scaling, uncovering power laws across various training and inference scaling techniques. We hope to inspire future work in scaling training and test-time compute for iterative generative paradigms such as diffusion for perception tasks.

\section{Acknowledgments}\label{sec:acknowledgments}
We thank Alexei Efros for helpful discussions. We also thank Xinlei Chen, Amil Dravid, Neerja Thakkar for their valuable feedback on the paper.

\bibliography{iclr2025_conference}

\begin{thebibliography}{56}
\providecommand{\natexlab}[1]{#1}
\providecommand{\url}[1]{\texttt{#1}}
\expandafter\ifx\csname urlstyle\endcsname\relax
  \providecommand{\doi}[1]{doi: #1}\else
  \providecommand{\doi}{doi: \begingroup \urlstyle{rm}\Url}\fi

\bibitem[Amit et~al.(2021)Amit, Shaharbany, Nachmani, and Wolf]{amit2021segdiff}
Tomer Amit, Tal Shaharbany, Eliya Nachmani, and Lior Wolf.
\newblock Segdiff: Image segmentation with diffusion probabilistic models.
\newblock \emph{arXiv preprint arXiv:2112.00390}, 2021.

\bibitem[Baranchuk et~al.(2021)Baranchuk, Rubachev, Voynov, Khrulkov, and Babenko]{baranchuk2021label}
Dmitry Baranchuk, Ivan Rubachev, Andrey Voynov, Valentin Khrulkov, and Artem Babenko.
\newblock Label-efficient semantic segmentation with diffusion models.
\newblock \emph{arXiv preprint arXiv:2112.03126}, 2021.

\bibitem[Brempong et~al.(2022)Brempong, Kornblith, Chen, Parmar, Minderer, and Norouzi]{brempong2022denoising}
Emmanuel~Asiedu Brempong, Simon Kornblith, Ting Chen, Niki Parmar, Matthias Minderer, and Mohammad Norouzi.
\newblock Denoising pretraining for semantic segmentation.
\newblock In \emph{Proceedings of the IEEE/CVF conference on computer vision and pattern recognition}, pp.\  4175--4186, 2022.

\bibitem[Brooks et~al.(2023)Brooks, Holynski, and Efros]{brooks2023instructpix2pixlearningfollowimage}
Tim Brooks, Aleksander Holynski, and Alexei~A. Efros.
\newblock Instructpix2pix: Learning to follow image editing instructions, 2023.
\newblock URL \url{https://arxiv.org/abs/2211.09800}.

\bibitem[Brown et~al.(2024)Brown, Juravsky, Ehrlich, Clark, Le, Ré, and Mirhoseini]{brown2024largelanguagemonkeysscaling}
Bradley Brown, Jordan Juravsky, Ryan Ehrlich, Ronald Clark, Quoc~V. Le, Christopher Ré, and Azalia Mirhoseini.
\newblock Large language monkeys: Scaling inference compute with repeated sampling, 2024.
\newblock URL \url{https://arxiv.org/abs/2407.21787}.

\bibitem[Chang et~al.(2023)Chang, Zhang, Barber, Maschinot, Lezama, Jiang, Yang, Murphy, Freeman, Rubinstein, Li, and Krishnan]{chang2023musetexttoimagegenerationmasked}
Huiwen Chang, Han Zhang, Jarred Barber, AJ~Maschinot, Jose Lezama, Lu~Jiang, Ming-Hsuan Yang, Kevin Murphy, William~T. Freeman, Michael Rubinstein, Yuanzhen Li, and Dilip Krishnan.
\newblock Muse: Text-to-image generation via masked generative transformers, 2023.
\newblock URL \url{https://arxiv.org/abs/2301.00704}.

\bibitem[Chen et~al.(2023)Chen, Li, Saxena, Hinton, and Fleet]{chen2023generalist}
Ting Chen, Lala Li, Saurabh Saxena, Geoffrey Hinton, and David~J Fleet.
\newblock A generalist framework for panoptic segmentation of images and videos.
\newblock In \emph{Proceedings of the IEEE/CVF international conference on computer vision}, pp.\  909--919, 2023.

\bibitem[Duan et~al.(2023)Duan, Guo, and Zhu]{duan2023diffusiondepth}
Yiqun Duan, Xianda Guo, and Zheng Zhu.
\newblock Diffusiondepth: Diffusion denoising approach for monocular depth estimation.
\newblock \emph{arXiv preprint arXiv:2303.05021}, 2023.

\bibitem[El-Refai et~al.(2024)El-Refai, Patel, Pei, and Li]{elswag}
Karim El-Refai, Zeeshan Patel, Jonathan Pei, and Tianle Li.
\newblock Swag: Storytelling with action guidance.
\newblock 2024.

\bibitem[Fei et~al.(2024{\natexlab{a}})Fei, Fan, Yu, Li, and Huang]{fei2024scaling}
Zhengcong Fei, Mingyuan Fan, Changqian Yu, Debang Li, and Junshi Huang.
\newblock Scaling diffusion transformers to 16 billion parameters.
\newblock \emph{arXiv preprint arXiv:2407.11633}, 2024{\natexlab{a}}.

\bibitem[Fei et~al.(2024{\natexlab{b}})Fei, Fan, Yu, Li, and Huang]{FeiDiTMoE2024}
Zhengcong Fei, Mingyuan Fan, Changqian Yu, Debang Li, and Jusnshi Huang.
\newblock Scaling diffusion transformers to 16 billion parameters.
\newblock \emph{arXiv preprint}, 2024{\natexlab{b}}.

\bibitem[Fischer et~al.(2015)Fischer, Dosovitskiy, Ilg, Häusser, Hazırbaş, Golkov, van~der Smagt, Cremers, and Brox]{fischer2015flownetlearningopticalflow}
Philipp Fischer, Alexey Dosovitskiy, Eddy Ilg, Philip Häusser, Caner Hazırbaş, Vladimir Golkov, Patrick van~der Smagt, Daniel Cremers, and Thomas Brox.
\newblock Flownet: Learning optical flow with convolutional networks, 2015.
\newblock URL \url{https://arxiv.org/abs/1504.06852}.

\bibitem[Fu et~al.(2024)Fu, Yin, Hu, Wang, Ma, Tan, Shen, Lin, and Long]{fu2024geowizardunleashingdiffusionpriors}
Xiao Fu, Wei Yin, Mu~Hu, Kaixuan Wang, Yuexin Ma, Ping Tan, Shaojie Shen, Dahua Lin, and Xiaoxiao Long.
\newblock Geowizard: Unleashing the diffusion priors for 3d geometry estimation from a single image, 2024.
\newblock URL \url{https://arxiv.org/abs/2403.12013}.

\bibitem[Goodfellow et~al.(2014)Goodfellow, Pouget-Abadie, Mirza, Xu, Warde-Farley, Ozair, Courville, and Bengio]{goodfellow2014generative}
Ian Goodfellow, Jean Pouget-Abadie, Mehdi Mirza, Bing Xu, David Warde-Farley, Sherjil Ozair, Aaron Courville, and Yoshua Bengio.
\newblock Generative adversarial nets.
\newblock \emph{Advances in neural information processing systems}, 27, 2014.

\bibitem[Gu et~al.(2024)Gu, Chen, and Xu]{gu2024diffusioninst}
Zhangxuan Gu, Haoxing Chen, and Zhuoer Xu.
\newblock Diffusioninst: Diffusion model for instance segmentation.
\newblock In \emph{ICASSP 2024-2024 IEEE International Conference on Acoustics, Speech and Signal Processing (ICASSP)}, pp.\  2730--2734. IEEE, 2024.

\bibitem[He et~al.(2024)He, Fu, Liu, Wang, Xiao, Shu, Wang, Zhang, Yu, Li, Huang, Gan, and Jiang]{he2024marsmixtureautoregressivemodels}
Wanggui He, Siming Fu, Mushui Liu, Xierui Wang, Wenyi Xiao, Fangxun Shu, Yi~Wang, Lei Zhang, Zhelun Yu, Haoyuan Li, Ziwei Huang, LeiLei Gan, and Hao Jiang.
\newblock Mars: Mixture of auto-regressive models for fine-grained text-to-image synthesis, 2024.
\newblock URL \url{https://arxiv.org/abs/2407.07614}.

\bibitem[Ho et~al.(2020)Ho, Jain, and Abbeel]{ho2020denoising}
Jonathan Ho, Ajay Jain, and Pieter Abbeel.
\newblock Denoising diffusion probabilistic models.
\newblock \emph{Advances in neural information processing systems}, 33:\penalty0 6840--6851, 2020.

\bibitem[Hoogeboom et~al.(2021)Hoogeboom, Nielsen, Jaini, Forr{\'e}, and Welling]{hoogeboom2021argmax}
Emiel Hoogeboom, Didrik Nielsen, Priyank Jaini, Patrick Forr{\'e}, and Max Welling.
\newblock Argmax flows and multinomial diffusion: Learning categorical distributions.
\newblock \emph{Advances in Neural Information Processing Systems}, 34:\penalty0 12454--12465, 2021.

\bibitem[Jain et~al.(2022)Jain, Mildenhall, Barron, Abbeel, and Poole]{jain2022zero}
Ajay Jain, Ben Mildenhall, Jonathan~T Barron, Pieter Abbeel, and Ben Poole.
\newblock Zero-shot text-guided object generation with dream fields.
\newblock In \emph{Proceedings of the IEEE/CVF conference on computer vision and pattern recognition}, pp.\  867--876, 2022.

\bibitem[Ji et~al.(2023)Ji, Chen, Xie, Hong, Liu, Liu, Lu, Li, and Luo]{ji2023ddp}
Yuanfeng Ji, Zhe Chen, Enze Xie, Lanqing Hong, Xihui Liu, Zhaoqiang Liu, Tong Lu, Zhenguo Li, and Ping Luo.
\newblock Ddp: Diffusion model for dense visual prediction.
\newblock In \emph{Proceedings of the IEEE/CVF International Conference on Computer Vision}, pp.\  21741--21752, 2023.

\bibitem[Ke et~al.(2024)Ke, Obukhov, Huang, Metzger, Daudt, and Schindler]{ke2024repurposing}
Bingxin Ke, Anton Obukhov, Shengyu Huang, Nando Metzger, Rodrigo~Caye Daudt, and Konrad Schindler.
\newblock Repurposing diffusion-based image generators for monocular depth estimation.
\newblock In \emph{Proceedings of the IEEE/CVF Conference on Computer Vision and Pattern Recognition}, pp.\  9492--9502, 2024.

\bibitem[Kingma(2013)]{kingma2013auto}
Diederik~P Kingma.
\newblock Auto-encoding variational bayes.
\newblock \emph{arXiv preprint arXiv:1312.6114}, 2013.

\bibitem[Komatsuzaki et~al.(2022)Komatsuzaki, Puigcerver, Lee-Thorp, Ruiz, Mustafa, Ainslie, Tay, Dehghani, and Houlsby]{komatsuzaki2022sparse}
Aran Komatsuzaki, Joan Puigcerver, James Lee-Thorp, Carlos~Riquelme Ruiz, Basil Mustafa, Joshua Ainslie, Yi~Tay, Mostafa Dehghani, and Neil Houlsby.
\newblock Sparse upcycling: Training mixture-of-experts from dense checkpoints.
\newblock \emph{arXiv preprint arXiv:2212.05055}, 2022.

\bibitem[Komatsuzaki et~al.(2023)Komatsuzaki, Puigcerver, Lee-Thorp, Ruiz, Mustafa, Ainslie, Tay, Dehghani, and Houlsby]{komatsuzaki2023sparseupcyclingtrainingmixtureofexperts}
Aran Komatsuzaki, Joan Puigcerver, James Lee-Thorp, Carlos~Riquelme Ruiz, Basil Mustafa, Joshua Ainslie, Yi~Tay, Mostafa Dehghani, and Neil Houlsby.
\newblock Sparse upcycling: Training mixture-of-experts from dense checkpoints, 2023.
\newblock URL \url{https://arxiv.org/abs/2212.05055}.

\bibitem[Li et~al.(2024)Li, Zou, Wang, Majumder, Xie, Manmatha, Swaminathan, Tu, Ermon, and Soatto]{li2024scalability}
Hao Li, Yang Zou, Ying Wang, Orchid Majumder, Yusheng Xie, R~Manmatha, Ashwin Swaminathan, Zhuowen Tu, Stefano Ermon, and Stefano Soatto.
\newblock On the scalability of diffusion-based text-to-image generation.
\newblock In \emph{Proceedings of the IEEE/CVF Conference on Computer Vision and Pattern Recognition}, pp.\  9400--9409, 2024.

\bibitem[Lipman et~al.(2023)]{lipman2023flowmatching}
Y~Lipman et~al.
\newblock Flow matching: Symmetrizing optimal transport and generative modeling.
\newblock \emph{arXiv preprint arXiv:2301.13003}, 2023.

\bibitem[Liu et~al.(2023)Liu, Wu, Van~Hoorick, Tokmakov, Zakharov, and Vondrick]{liu2023zero}
Ruoshi Liu, Rundi Wu, Basile Van~Hoorick, Pavel Tokmakov, Sergey Zakharov, and Carl Vondrick.
\newblock Zero-1-to-3: Zero-shot one image to 3d object.
\newblock In \emph{Proceedings of the IEEE/CVF international conference on computer vision}, pp.\  9298--9309, 2023.

\bibitem[Liu et~al.(2022)]{liu2022flowmatching}
X~Liu et~al.
\newblock Rectified flow: A unified approach for free-form generative models.
\newblock \emph{arXiv preprint arXiv:2209.07953}, 2022.

\bibitem[Luo et~al.(2024)Luo, Li, Yang, Liu, Fan, and Liu]{luo2024flowdiffuser}
Ao~Luo, Xin Li, Fan Yang, Jiangyu Liu, Haoqiang Fan, and Shuaicheng Liu.
\newblock Flowdiffuser: Advancing optical flow estimation with diffusion models.
\newblock In \emph{Proceedings of the IEEE/CVF Conference on Computer Vision and Pattern Recognition}, pp.\  19167--19176, 2024.

\bibitem[Nichol \& Dhariwal(2021)Nichol and Dhariwal]{nichol2021improveddenoisingdiffusionprobabilistic}
Alex Nichol and Prafulla Dhariwal.
\newblock Improved denoising diffusion probabilistic models, 2021.
\newblock URL \url{https://arxiv.org/abs/2102.09672}.

\bibitem[OpenAI(2024)]{openai2024o1}
OpenAI.
\newblock Learning to reason with llms.
\newblock \url{https://openai.com/index/learning-to-reason-with-llms/}, September 2024.

\bibitem[Ozguroglu et~al.(2024)Ozguroglu, Liu, Sur\'s, Chen, Dave, Tokmakov, and Vondrick]{ozguroglu2024pix2gestalt}
Ege Ozguroglu, Ruoshi Liu, D\'idac Sur\'s, Dian Chen, Achal Dave, Pavel Tokmakov, and Carl Vondrick.
\newblock pix2gestalt: Amodal segmentation by synthesizing wholes.
\newblock \emph{Proceedings of the IEEE/CVF Conference on Computer Vision and Pattern Recognition (CVPR)}, 2024.

\bibitem[Peebles \& Xie(2023)Peebles and Xie]{peebles2023scalable}
William Peebles and Saining Xie.
\newblock Scalable diffusion models with transformers.
\newblock In \emph{Proceedings of the IEEE/CVF International Conference on Computer Vision}, pp.\  4195--4205, 2023.

\bibitem[Poole et~al.(2022)Poole, Jain, Barron, and Mildenhall]{poole2022dreamfusion}
Ben Poole, Ajay Jain, Jonathan~T Barron, and Ben Mildenhall.
\newblock Dreamfusion: Text-to-3d using 2d diffusion.
\newblock \emph{arXiv preprint arXiv:2209.14988}, 2022.

\bibitem[Rezende \& Mohamed(2015)Rezende and Mohamed]{rezende2015variational}
Danilo Rezende and Shakir Mohamed.
\newblock Variational inference with normalizing flows.
\newblock In \emph{International conference on machine learning}, pp.\  1530--1538. PMLR, 2015.

\bibitem[Rombach et~al.(2022)Rombach, Blattmann, Lorenz, Esser, and Ommer]{rombach2022high}
Robin Rombach, Andreas Blattmann, Dominik Lorenz, Patrick Esser, and Bj{\"o}rn Ommer.
\newblock High-resolution image synthesis with latent diffusion models.
\newblock In \emph{Proceedings of the IEEE/CVF conference on computer vision and pattern recognition}, pp.\  10684--10695, 2022.

\bibitem[Russakovsky et~al.(2015)Russakovsky, Deng, Su, Krause, Satheesh, Ma, Huang, Karpathy, Khosla, Bernstein, Berg, and Fei-Fei]{russakovsky2015imagenetlargescalevisual}
Olga Russakovsky, Jia Deng, Hao Su, Jonathan Krause, Sanjeev Satheesh, Sean Ma, Zhiheng Huang, Andrej Karpathy, Aditya Khosla, Michael Bernstein, Alexander~C. Berg, and Li~Fei-Fei.
\newblock Imagenet large scale visual recognition challenge, 2015.
\newblock URL \url{https://arxiv.org/abs/1409.0575}.

\bibitem[Saharia et~al.(2022)Saharia, Chan, Saxena, Li, Whang, Denton, Ghasemipour, Gontijo~Lopes, Karagol~Ayan, Salimans, et~al.]{saharia2022photorealistic}
Chitwan Saharia, William Chan, Saurabh Saxena, Lala Li, Jay Whang, Emily~L Denton, Kamyar Ghasemipour, Raphael Gontijo~Lopes, Burcu Karagol~Ayan, Tim Salimans, et~al.
\newblock Photorealistic text-to-image diffusion models with deep language understanding.
\newblock \emph{Advances in neural information processing systems}, 35:\penalty0 36479--36494, 2022.

\bibitem[Saxena et~al.(2023)Saxena, Kar, Norouzi, and Fleet]{saxena2023monocular}
Saurabh Saxena, Abhishek Kar, Mohammad Norouzi, and David~J Fleet.
\newblock Monocular depth estimation using diffusion models.
\newblock \emph{arXiv preprint arXiv:2302.14816}, 2023.

\bibitem[Saxena et~al.(2024)Saxena, Herrmann, Hur, Kar, Norouzi, Sun, and Fleet]{saxena2024surprising}
Saurabh Saxena, Charles Herrmann, Junhwa Hur, Abhishek Kar, Mohammad Norouzi, Deqing Sun, and David~J Fleet.
\newblock The surprising effectiveness of diffusion models for optical flow and monocular depth estimation.
\newblock \emph{Advances in Neural Information Processing Systems}, 36, 2024.

\bibitem[Shazeer et~al.(2017)Shazeer, Mirhoseini, Maziarz, Davis, Le, Hinton, and Dean]{shazeer2017outrageouslylargeneuralnetworks}
Noam Shazeer, Azalia Mirhoseini, Krzysztof Maziarz, Andy Davis, Quoc Le, Geoffrey Hinton, and Jeff Dean.
\newblock Outrageously large neural networks: The sparsely-gated mixture-of-experts layer, 2017.
\newblock URL \url{https://arxiv.org/abs/1701.06538}.

\bibitem[Snell et~al.(2024)Snell, Lee, Xu, and Kumar]{snell2024scalingllmtesttimecompute}
Charlie Snell, Jaehoon Lee, Kelvin Xu, and Aviral Kumar.
\newblock Scaling llm test-time compute optimally can be more effective than scaling model parameters, 2024.
\newblock URL \url{https://arxiv.org/abs/2408.03314}.

\bibitem[Sohl-Dickstein et~al.(2015)Sohl-Dickstein, Weiss, Maheswaranathan, and Ganguli]{sohl2015deep}
Jascha Sohl-Dickstein, Eric Weiss, Niru Maheswaranathan, and Surya Ganguli.
\newblock Deep unsupervised learning using nonequilibrium thermodynamics.
\newblock In \emph{International conference on machine learning}, pp.\  2256--2265. PMLR, 2015.

\bibitem[Song et~al.(2021)Song, Meng, and Ermon]{song2021denoising}
Jiaming Song, Chenlin Meng, and Stefano Ermon.
\newblock Denoising diffusion implicit models.
\newblock In \emph{International Conference on Learning Representations}, 2021.
\newblock URL \url{https://openreview.net/forum?id=St1giarCHLP}.

\bibitem[Song et~al.(2023)Song, Dhariwal, Chen, and Sutskever]{song2023consistency}
Yang Song, Prafulla Dhariwal, Mark Chen, and Ilya Sutskever.
\newblock Consistency models.
\newblock \emph{arXiv preprint arXiv:2303.01469}, 2023.

\bibitem[Sun et~al.(2024)Sun, Lei, Zhang, Li, Huang, Pang, Dai, and Du]{sun2024ecditscalingdiffusiontransformers}
Haotian Sun, Tao Lei, Bowen Zhang, Yanghao Li, Haoshuo Huang, Ruoming Pang, Bo~Dai, and Nan Du.
\newblock Ec-dit: Scaling diffusion transformers with adaptive expert-choice routing, 2024.
\newblock URL \url{https://arxiv.org/abs/2410.02098}.

\bibitem[Tan et~al.(2022)Tan, Wu, and Pi]{tan2022semantic}
Haoru Tan, Sitong Wu, and Jimin Pi.
\newblock Semantic diffusion network for semantic segmentation.
\newblock \emph{Advances in Neural Information Processing Systems}, 35:\penalty0 8702--8716, 2022.

\bibitem[Touvron et~al.(2023)Touvron, Martin, Stone, Albert, Almahairi, Babaei, Bashlykov, Batra, Bhargava, Bhosale, Bikel, Blecher, Ferrer, Chen, Cucurull, Esiobu, Fernandes, Fu, Fu, Fuller, Gao, Goswami, Goyal, Hartshorn, Hosseini, Hou, Inan, Kardas, Kerkez, Khabsa, Kloumann, Korenev, Koura, Lachaux, Lavril, Lee, Liskovich, Lu, Mao, Martinet, Mihaylov, Mishra, Molybog, Nie, Poulton, Reizenstein, Rungta, Saladi, Schelten, Silva, Smith, Subramanian, Tan, Tang, Taylor, Williams, Kuan, Xu, Yan, Zarov, Zhang, Fan, Kambadur, Narang, Rodriguez, Stojnic, Edunov, and Scialom]{touvron2023llama2openfoundation}
Hugo Touvron, Louis Martin, Kevin Stone, Peter Albert, Amjad Almahairi, Yasmine Babaei, Nikolay Bashlykov, Soumya Batra, Prajjwal Bhargava, Shruti Bhosale, Dan Bikel, Lukas Blecher, Cristian~Canton Ferrer, Moya Chen, Guillem Cucurull, David Esiobu, Jude Fernandes, Jeremy Fu, Wenyin Fu, Brian Fuller, Cynthia Gao, Vedanuj Goswami, Naman Goyal, Anthony Hartshorn, Saghar Hosseini, Rui Hou, Hakan Inan, Marcin Kardas, Viktor Kerkez, Madian Khabsa, Isabel Kloumann, Artem Korenev, Punit~Singh Koura, Marie-Anne Lachaux, Thibaut Lavril, Jenya Lee, Diana Liskovich, Yinghai Lu, Yuning Mao, Xavier Martinet, Todor Mihaylov, Pushkar Mishra, Igor Molybog, Yixin Nie, Andrew Poulton, Jeremy Reizenstein, Rashi Rungta, Kalyan Saladi, Alan Schelten, Ruan Silva, Eric~Michael Smith, Ranjan Subramanian, Xiaoqing~Ellen Tan, Binh Tang, Ross Taylor, Adina Williams, Jian~Xiang Kuan, Puxin Xu, Zheng Yan, Iliyan Zarov, Yuchen Zhang, Angela Fan, Melanie Kambadur, Sharan Narang, Aurelien Rodriguez, Robert Stojnic, Sergey Edunov, and Thomas
  Scialom.
\newblock Llama 2: Open foundation and fine-tuned chat models, 2023.
\newblock URL \url{https://arxiv.org/abs/2307.09288}.

\bibitem[van~den Oord et~al.(2016)van~den Oord, Kalchbrenner, and Kavukcuoglu]{pmlr-v48-oord16}
Aäron van~den Oord, Nal Kalchbrenner, and Koray Kavukcuoglu.
\newblock Pixel recurrent neural networks.
\newblock In Maria~Florina Balcan and Kilian~Q. Weinberger (eds.), \emph{Proceedings of The 33rd International Conference on Machine Learning}, volume~48 of \emph{Proceedings of Machine Learning Research}, pp.\  1747--1756, New York, New York, USA, 20--22 Jun 2016. PMLR.
\newblock URL \url{https://proceedings.mlr.press/v48/oord16.html}.

\bibitem[Wang et~al.(2023)Wang, Du, Li, Yeh, and Shakhnarovich]{wang2023score}
Haochen Wang, Xiaodan Du, Jiahao Li, Raymond~A Yeh, and Greg Shakhnarovich.
\newblock Score jacobian chaining: Lifting pretrained 2d diffusion models for 3d generation.
\newblock In \emph{Proceedings of the IEEE/CVF Conference on Computer Vision and Pattern Recognition}, pp.\  12619--12629, 2023.

\bibitem[Watson et~al.(2022)Watson, Chan, Martin-Brualla, Ho, Tagliasacchi, and Norouzi]{watson2022novel}
Daniel Watson, William Chan, Ricardo Martin-Brualla, Jonathan Ho, Andrea Tagliasacchi, and Mohammad Norouzi.
\newblock Novel view synthesis with diffusion models.
\newblock \emph{arXiv preprint arXiv:2210.04628}, 2022.

\bibitem[Weinzaepfel et~al.(2013)Weinzaepfel, Revaud, Harchaoui, and Schmid]{Weinzaepfel_2013_ICCV}
Philippe Weinzaepfel, Jerome Revaud, Zaid Harchaoui, and Cordelia Schmid.
\newblock Deepflow: Large displacement optical flow with deep matching.
\newblock In \emph{Proceedings of the IEEE International Conference on Computer Vision (ICCV)}, December 2013.

\bibitem[Wolleb et~al.(2022)Wolleb, Sandk{\"u}hler, Bieder, Valmaggia, and Cattin]{wolleb2022diffusion}
Julia Wolleb, Robin Sandk{\"u}hler, Florentin Bieder, Philippe Valmaggia, and Philippe~C Cattin.
\newblock Diffusion models for implicit image segmentation ensembles.
\newblock In \emph{International Conference on Medical Imaging with Deep Learning}, pp.\  1336--1348. PMLR, 2022.

\bibitem[Yang et~al.(2022)Yang, Hu, Babuschkin, Sidor, Liu, Farhi, Ryder, Pachocki, Chen, and Gao]{yang2022tensorprogramsvtuning}
Greg Yang, Edward~J. Hu, Igor Babuschkin, Szymon Sidor, Xiaodong Liu, David Farhi, Nick Ryder, Jakub Pachocki, Weizhu Chen, and Jianfeng Gao.
\newblock Tensor programs v: Tuning large neural networks via zero-shot hyperparameter transfer, 2022.
\newblock URL \url{https://arxiv.org/abs/2203.03466}.

\bibitem[Yu et~al.(2022)Yu, Xu, Koh, Luong, Baid, Wang, Vasudevan, Ku, Yang, Ayan, et~al.]{yu2022scaling}
Jiahui Yu, Yuanzhong Xu, Jing~Yu Koh, Thang Luong, Gunjan Baid, Zirui Wang, Vijay Vasudevan, Alexander Ku, Yinfei Yang, Burcu~Karagol Ayan, et~al.
\newblock Scaling autoregressive models for content-rich text-to-image generation.
\newblock \emph{arXiv preprint arXiv:2206.10789}, 2\penalty0 (3):\penalty0 5, 2022.

\bibitem[Zhao et~al.(2023)Zhao, Rao, Liu, Liu, Zhou, and Lu]{zhao2023unleashing}
Wenliang Zhao, Yongming Rao, Zuyan Liu, Benlin Liu, Jie Zhou, and Jiwen Lu.
\newblock Unleashing text-to-image diffusion models for visual perception.
\newblock In \emph{Proceedings of the IEEE/CVF International Conference on Computer Vision}, pp.\  5729--5739, 2023.

\end{thebibliography}
\bibliographystyle{iclr2025_conference}

\end{document}